\documentclass[10pt,twocolumn,letterpaper]{article}

\usepackage{iccv}
\usepackage{times}
\usepackage{epsfig}
\usepackage{graphicx}
\usepackage{amssymb,amsmath,amsthm}
\usepackage{algorithmic}
\usepackage[ruled,vlined]{algorithm2e}
\usepackage{subfigure}
\usepackage[dvipsnames]{xcolor}
\usepackage{enumitem}
\usepackage[accsupp]{axessibility}
\usepackage{etoc}
\newtheorem{theorem}{Theorem}[section]
\newtheorem{lemma}[theorem]{Lemma}

\usepackage[toc,page,header]{appendix}
\usepackage{minitoc}

\usepackage[pagebackref=true,breaklinks=true,letterpaper=true,colorlinks,bookmarks=false]{hyperref}
\iccvfinalcopy %

\def\iccvPaperID{10567} %

\begin{document}

\title{S$^3$VAADA: Submodular Subset Selection for Virtual Adversarial Active Domain Adaptation }

\author{Harsh Rangwani \quad Arihant Jain\thanks{Equal Contribution} \quad Sumukh K Aithal\footnotemark[1] \quad R. Venkatesh Babu\\
Video Analytics Lab, Indian Institute of Science, Bengaluru, India \\
{\tt\small harshr@iisc.ac.in, arihantjain@iisc.ac.in, sumukhaithal6@gmail.com, venky@iisc.ac.in}

}

\maketitle
\ificcvfinal\thispagestyle{empty}\fi

\begin{abstract}
   Unsupervised domain adaptation (DA) methods have focused on achieving maximal performance through aligning features from source and target domains without using labeled data in the target domain. Whereas, in the real-world scenario’s it might be feasible to get labels for a small proportion of target data. In these scenarios, it is important to select maximally-informative samples to label and find an effective way to combine them with the existing knowledge from source data. Towards achieving this, we propose S$^3$VAADA which i) introduces a novel submodular criterion to select a maximally informative subset to label and ii) enhances a cluster-based DA procedure through novel improvements to effectively utilize all the available data for improving generalization on target. Our approach consistently outperforms the competing state-of-the-art approaches on datasets with varying degrees of domain shifts. The project page with additional details is available here: \href{https://sites.google.com/iisc.ac.in/s3vaada-iccv2021/}{https://sites.google.com/iisc.ac.in/s3vaada-iccv2021/}.

\end{abstract}

\section{Introduction}

Deep Neural Networks have shown significant advances in image classification tasks by utilizing large amounts of labeled data. Despite their impressive performance, these networks produce spurious predictions hence suffer from performance degradation when used on images that come from a different domain~\cite{saenko2010adapting} (e.g., model trained on synthetic data (source domain) being used on real-world data (target domain)). Unsupervised Domain Adaptation (DA)~\cite{ganin2015unsupervised, long2018conditional, chen2020adversarial, saito2018adversarial,Kundu_2020_CVPR, Kundu_2020_CVPR_usfda} approaches aim to utilize the labeled data from source domain along with the unlabeled data from the target domain to improve the model’s generalization on the target domain. However, it has been observed that the performance of Unsupervised DA models often falls short in comparison to the supervised methods \cite{tsai2018learning}, which leads to their reduced usage for performance critical applications. In such cases, it might be possible to label some of the target data to improve the performance of the model. 
\begin{figure}[t]
    \centering
    \includegraphics[width=0.45\textwidth]{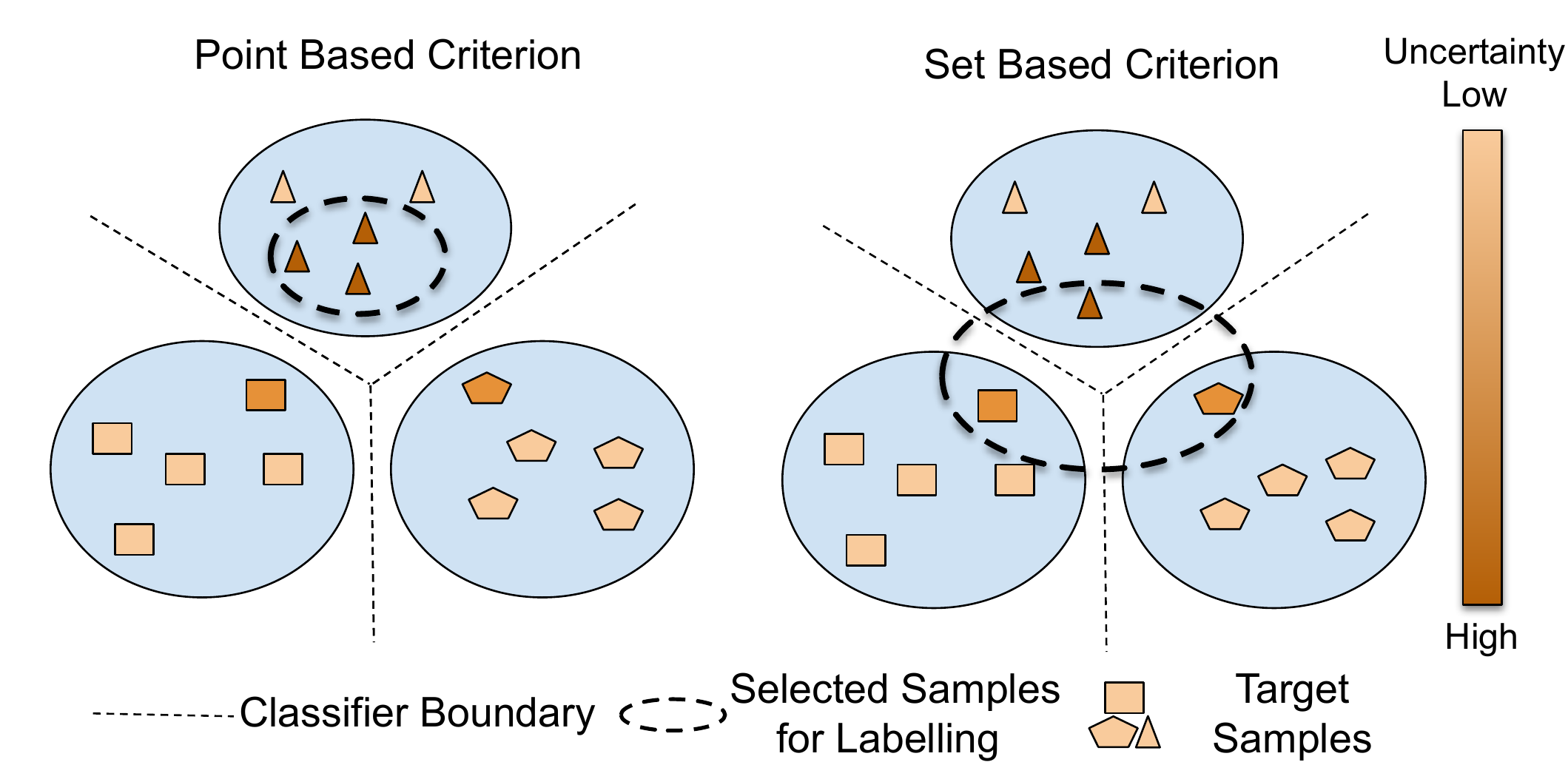}
    \caption{We pose sample selection  for labeling in \textbf{Active Domain Adaptation} as an informative subset selection problem. We propose an information criterion to provide \textit{score for each subset} of samples for labeling. Prior works (See t-SNE for AADA~\cite{Su_2020_WACV} in Sec. 1 of supplementary material) which use a point-based criterion \textit{(i.e. score each sample independently)} to select samples suffer from redundancy. As our set-based criterion is aware of the other samples in the set, it avoids redundancy and selects diverse samples. }
    \label{fig:inuitive_explanation}
\end{figure}

\begin{figure*}[htp]
  \centering
  \includegraphics[width=\linewidth]{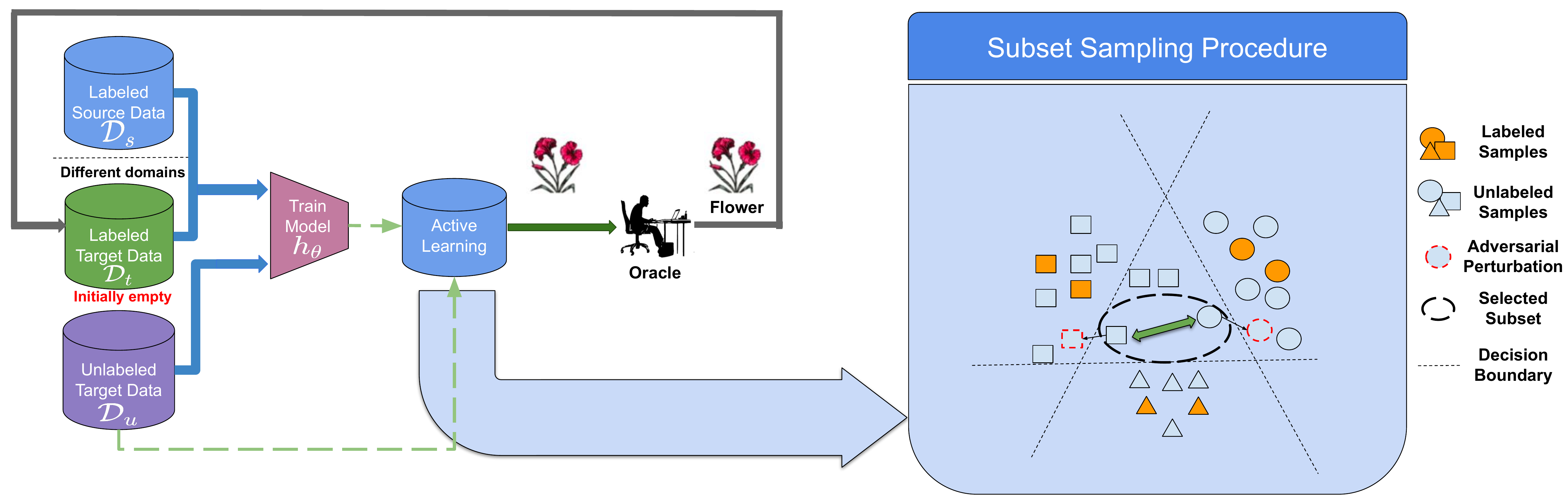}
  \caption{Overview of Submodular Subset Selection for Virtual Adversarial Active Adaptation (S$^3$VAADA). Step 1: We select a subset of samples which are uncertain (i.e., prediction can change with small perturbation), diverse and representative in feature space (see Fig. \ref{fig:sampling-al} for details). Step 2: The labeled samples and the unlabeled samples are used by proposed VAADA adaptation procedure to obtain the final model. The above two steps are iteratively followed selecting $B$ samples in each cycle, till the annotation budget is exhausted.}
  \label{fig:al}
\end{figure*}

In such a case, the dilemma is, \textit{“Which samples from the target dataset should be selected for labeling?”}. Active Learning (AL)~\cite{cohn1994improving, settles2009active} approaches aim to provide techniques to select the maximally informative set for labeling, which is then used for training the model. However, these approaches do not effectively use the unlabeled data and labeled data present in various domains. This objective contrasts with Unsupervised DA objective that aims to use the unlabeled target data effectively. In practice, it has been found that just naively using AL and fine-tuning offers sub-optimal performance in presence of domain shift~\cite{Su_2020_WACV}. 

Another question that follows sample selection (or sampling) is, \textit{“How to effectively use all the data available to improve model performance?”}. Unsupervised DA approaches based on the idea of learning invariant features for both the source and target domain have been known to be ineffective in increasing performance when additional labeled data is present in target domain~\cite{saito2019semi}. Semi-Supervised DA (SSDA)~\cite{saito2019semi, li2020online} methods have been developed to mitigate the above issue, but we find their performance plateau’s as additional data is added (Sec. \ref{sec:results}). This is likely due to the assumption in SSDA of only having a small amount of labeled data per-class (i.e., few shot) in target domain which is restrictive.

The Active Domain Adaptation (Active DA) paradigm introduced by Rai et al. \cite{rai2010domain} aims to first effectively select the informative-samples, which are then used by a complementary DA procedure for improving model performance on target domain. The state-of-the-art work of AADA~\cite{Su_2020_WACV} aim to select samples with high value of $p_{target}(x)/p_{source}(x)$ from domain discriminator, multiplied by the entropy of classifier which is used by DANN \cite{ganin2015unsupervised} for adaptation. As the AADA criterion is a point-estimate, it is unaware of other selected samples; hence the samples selected can be redundant, as shown in Fig. \ref{fig:inuitive_explanation}.

In this work, we introduce  Submodular Subset Selection for Virtual Adversarial Active Domain Adaptation (S$^3$VAADA) which proposes a set-based informative criterion that provides scores for each of the subset of samples rather than a point-based estimate for each sample. As the information criteria is aware of other samples in the subset, it tends to avoid redundancy. Hence, it is able to select diverse and uncertain samples (shown in Fig. \ref{fig:inuitive_explanation}). Our subset criterion is based on the cluster assumption, which has shown to be widely effective in DA ~\cite{deng2019cluster, lee2019drop}. The subset criterion is composed of a novel uncertainty score (Virtual Adversarial Pairwise (VAP)) which is based on the idea of the sensitivity of model to small adversarial perturbations. This is combined with a distance based metrics such that the criterion is submodular (defined in Sec. \ref{sec:notations}). The submodularity of the criterion allows usage of an efficient algorithm \cite{nemhauser1978analysis} to obtain the optimal subset of samples.  After obtaining the labeled data, we use a cluster based domain adaption scheme based on VADA \cite{shu2018dirt}. Although VADA, when naively used is not able to effectively make use of the additional target labeled data \cite{saito2018maximum}, we mitigate this via two modifications (Sec. \ref{section:VAADA}) which form our Virtual Active Adversarial Domain Adaptation (VAADA) procedure. \\
In summary, our contributions are: 
\vspace{-1mm}
\begin{itemize}[noitemsep]
\item We propose a novel set-based information criterion which is aware of other samples in the set and aims to select uncertain, diverse and representative samples.
\item For effective utilization of the selected samples, we propose a complementary DA procedure of VAADA which enhances VADA's suitability for active DA.%
\item Our method demonstrates state-of-the-art active DA results on diverse domain adaptation benchmarks of Office-31, Office-Home and VisDA-18.
\end{itemize}

\section{Related Work}
\label{sec:rel_work}
\textbf{Domain Adaptation}:  One of the central ideas in DA is minimizing the discrepancy in two domains by aligning them in feature space. 
DANN~\cite{ganin2015unsupervised} achieves this by using domain classifier which is trained through an adversarial min-max objective to align the features of source and target domain. MCD~\cite{saito2018maximum} tries to minimize the discrepancy by making use of two classifiers trained in an adversarial fashion for aligning the features in two domains. The idea of semi-supervised domain adaptation by using a fraction of labeled data is also introduced in MME~\cite{saito2019semi} approach, which induces the feature invariance by a MinMax Entropy objective. Another set of approaches uses the cluster assumption to cluster the samples of the target domain and source domain. In our work, we use ideas from VADA (Virtual Adversarial Domain Adaptation)~\cite{shu2018dirt} to enforce cluster assumption. \\
\textbf{Active Learning (AL):} The traditional AL methods are iterative schemes which obtain labels (from oracle or experts) for a set of informative data samples. The newly labeled samples are then added to the pool of existing labeled data and the model is trained again on the labeled data. The proposed techniques can be divided into two classes: 1) \textbf{Uncertainty Based Methods}: In this case, model uncertainty about a particular sample is measured by specific criterion like entropy~\cite{wang2014new} etc. 2) \textbf{Diversity or Coverage Based Methods}: These methods focus on selecting a diverse set of points to label in order to improve the overall performance of the model. One of the popular methods, in this case, is Core-Set~\cite{sener2018active} which selects samples to maximize the coverage in feature space. However, recent approaches like BADGE~\cite{Ash2020Deep} which use a combination of uncertainty and diversity, achieve state-of-the-art performance.
A few task-agnostic active learning methods \cite{sinha2019variational, yoo2019learning} have also been proposed.\\\\
\textbf{Active Domain Adaptation:} The first attempt for active domain adaptation was made by Rai et al. \cite{rai2010domain}, who use linear classifier based criteria to select samples to label for sentiment analysis. Chattopadhyay et al. \cite{chattopadhyay2013joint} proposed a method to perform domain adaptation and active learning by solving a single convex optimization problem. AADA (Active Adversarial Domain Adaptation)~\cite{Su_2020_WACV} for image based DA is a method which proposes a hybrid informativeness criterion based on the output of classifier and domain discriminator used in DANN. The criterion used in AADA for selecting a batch used is a point estimate, which might lead to redundant sample selection. We introduce a set-based informativeness criterion to select samples to be labeled. CLUE~\cite{prabhu2020active} is a recent concurrent work which selects samples through uncertainty-weighted clustering for Active DA.

\begin{figure*}[!t]
  \centering
  \includegraphics[width=\linewidth]{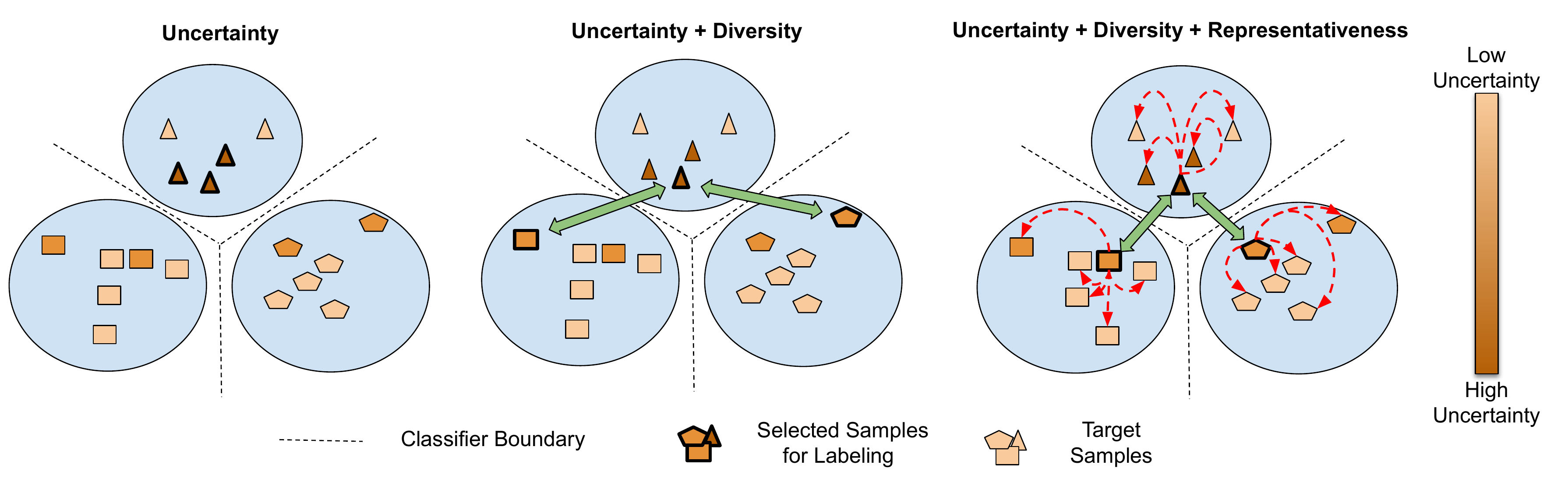}
  \caption{Our sampling technique incorporates uncertainty, diversity and representativeness. Just using {\color{brown} uncertainty} can lead to \textit{redundant} sample selection as shown on left. Whereas, incorporating diversity to ensure a large {\color{YellowGreen} distance} between selected samples may lead to selection of \textit{outliers}. Our sampling technique avoids outliers by selecting uncertain samples which are {\color{red} representative} of the clusters.}
  \label{fig:sampling-al}
\end{figure*}
\label{sec:S^3}

\section{Background}
\subsection{Definitions and Notations}
\label{sec:notations}
\textbf{Definitions:} We first define a set function $f(S)$ for which input is a set $S$. A submodular function is a set function $f:2^{\Omega}\rightarrow \mathbb{R}$, where $2^{\Omega}$ is the power set of set $\Omega$ which contains all elements. The submodular functions are characterized by the property of diminishing returns i.e., addition of a new element to smaller set must produce a larger increase in $f$ in comparison to addition to a larger set. This is mathematically stated as for every $S_1,S_2 \subseteq \Omega$ having $S_1 \subseteq S_2$ then for every $x \in \Omega \backslash S_2$ the following property holds:
\begin{equation}
\begin{split}
    f(S_1 \cup \{x\}) - f(S_1)  \geq  f(S_2\cup \{x\}) - f(S_2) %
\end{split}
\end{equation}
This property is known as the \textit{diminishing returns} property. 
\\
\textbf{Notations Used:} In the subsequent sections we use $h_{\theta}(x)$ as softmax output of the classifier, $h_{\theta}(x)$ is a composition of $f_{\theta} \circ g_{\theta} (x)$ where, $g_{\theta}(x)$ is the function that maps input to embedding and $f_{\theta}$ does final classification. The domain discriminator is a network $D_{\phi}(g_{\theta}(x))$ which classifies the sample into source and target domain which adversarially aligns the domains. We use $\mathcal{D}$ for combined data from both domains and use symbols of $\mathcal{D}_s$ and $\mathcal{D}_t$ for labeled data from source and target domain respectively. $\mathcal{D}_u$ denotes the unlabeled target data. In active DA, we define budget $B$ as number of target samples selected from $\mathcal{D}_u$ and added to $\mathcal{D}_t$ in each cycle.
\\
\textbf{Active Domain Adaptation:}
In each cycle, we first perform DA using $\mathcal{D}_s$ and $\mathcal{D}_t$ as the source and $\mathcal{D}_u$ as the target. Active Learning techniques are then utilized to select $B$ most informative samples from $\mathcal{D}_u$ which is then added to $\mathcal{D}_t$. This is performed for $C$ cycles.

\subsection{Cluster Assumption}
\label{sec:cluster}
Cluster assumption states that the decision boundaries should not lie in high density regions of data samples, which is a prime motivation for our approach. For enforcing cluster assumption we make use of two additional objectives from VADA ~\cite{shu2018dirt} method. The first objective is the minimization of conditional entropy on the unlabeled target data $\mathcal{D}_u$. This is enforced by using the following loss function:
\begin{equation}
    \label{eq:Conditional Entropy}
    L_c(\theta; \mathcal{D}_u) = - \mathbb{E}_{x \sim \mathcal{D}_u}[h_{\theta}(x)^T\ln{  h_{\theta}(x)}]
\end{equation}
The above objective ensures the formation of clusters of target samples, as it enforces high-confidence for classification on target data. 
However due to large capacity of neural networks, the classification function learnt can be locally non Lipschitz which can allow for large change in function value with small change in input. This leads to unreliable estimates of the conditional entropy loss $L_c$. For enforcing the local Lipschitzness we use the second objective, which was originally proposed in Virtual Adversarial Training (VAT) ~\cite{miyato2019vat}. It ensures smoothness in the $\epsilon$ norm ball enclosing the samples. The VAT objective is given below: 
\begin{equation}
    \label{eq: Vat}
    L_{v}(\theta; \mathcal{D}) = \mathbb{E}_{x \sim \mathcal{D}}[\underset{||r||\leq \epsilon}\max{D_{KL}(h_{\theta}(x) || h_{\theta}(x + r))}]
\end{equation}
\section{Proposed Method}
In Active Domain Adaptation, there are two distinct steps i.e., sample selection (i.e. sampling) followed by Domain Adaptation which we describe below: 
\subsection{Submodular Subset Selection}

\subsubsection{Virtual Adversarial Pairwise (VAP) Score}
\label{sec:vapscore}
In our model architecture, we only use a linear classifier and a softmax over domain invariant features $f_\theta(x)$ for classification. Due to the linear nature, we draw inspiration from SVM (Support Vector Machines) based AL methods which demonstrate that the samples near the boundary are likely to be support vectors, hence are more informative than other samples. There is also the theoretical justification behind choosing samples which are near the boundary in case of SVMs \cite{tong2001support}.
Hence we also aim to find the vectors which are near to the boundary by adversarially perturbing each sample $x$. We use the following objective to create perturbation:
\begin{equation}
    \underset{||r_i||\leq \epsilon}{max} D_{KL}(h_{\theta}(x) || h_{\theta}(x + r_i))
\end{equation}

Power Method \cite{Goluba2000EigenvalueCI} is used for finding the adversarial perturbation $r_i$ which involves initialization of random vector. As we aim to find vectors which are near the decision boundary, there can be cases where a particular sample may lie close to multiple decision boundaries as we operate in the setting of multi-class classification. Hence we create the perturbation $r_i$ for $N$ number of random initializations. This is done to select samples which can be easily perturbed to a diverse set of classes and also increase the reliability of the uncertainty estimate. We use the mean pairwise KL divergence score of probability distribution as a metric for measuring the uncertainty of sample. This is defined as Virtual Adversarial Pairwise (VAP) score given formally as:
\begin{equation} 
    \begin{split}
            VAP(x) = \frac{1}{N^2} \Bigg( \sum_{i = 1}^{N} D_{KL}(h_{\theta}(x) || h_{\theta}(x + r_i)) \\ +\ \sum_{i=1}^{N}\sum_{j=1, i \neq j}^{N} D_{KL}(h_{\theta}(x + r_i) || h_{\theta}(x + r_j)) \Bigg) 
    \end{split}
    \label{eq:vap score}
\end{equation}
The first term corresponds to divergence between perturbed input and the original sample $x$, the second term corresponds to diversity in the output of different perturbations. The approach is pictorially depicted on the right side in Fig. \ref{fig:al}. For VAP score to be meaningful we assume that cluster assumption holds and the function is smooth, which makes VAADA a complementary DA approach to our method. 
\subsubsection{Diversity Score}
\vspace{-2mm}
Just using VAP score for sampling can suffer from the issue of multiple similar samples being selected from the same cluster. For selecting the diverse samples in our set $S$ we propose to use the following diversity score for sample $x_i$ which is not present in $S$.
\begin{equation}
    d(S, x_i) = \underset{x \in S }{\min}D(x, x_i)
\end{equation}
where $D$ is a function of divergence. In our case we use the KL Divergence function:
\begin{equation}
    D(x_j, x_i) = D_{KL}(h_{\theta}(x_j)|| h_{\theta}(x_i)) 
\end{equation}
\subsubsection{Representativeness Score}
The combination of above two scores can ensure that the diverse and uncertain samples are selected. But this could still lead to selection of outliers as they can also be uncertain and diversely placed in feature space. For mitigating this we use a term based on facility location problem \cite{enwiki:1012600046} which ensures that selected samples are placed such that they are representative of unlabeled set. The score is mathematically defined as:
\begin{equation}
R(S, x_i) = \underset{x_k \in \mathcal{D}_u}{\sum} \max(0, s_{ki} - \underset{x_j \in S}{\max} \; s_{kj}) 
\end{equation}
The $s_{ij}$ corresponds to the similarity between sample $x_i$ and $x_j$. We use the similarity function $-\ln(1 - BC(h_{\theta}(x_i),h_{\theta}(x_j))$ where $BC(p,q)$ is the Bhattacharya coefficient~\cite{10.2307/25047882} defined as $\sum_{k} \sqrt{p_kq_k}$ for probability distributions $p$ and $q$.

\subsubsection{Combining the Three Score Functions}
\label{sec: Combinaton}

We define the set function $f(S)$ by defining the gain as a convex combination of $VAP(x_i)$ , $d(S,x_i)$ and $R(S,x_i)$.
\begin{equation}
    \begin{split}
    \label{eq:submod}
     f(S \cup \{x_i\}) - f(S) = \alpha VAP(x_i) + \beta d(S, x_i) \\
     +\ (1 - \alpha - \beta) R(S,x_i)
\end{split}
\end{equation}

Here $0\leq \alpha, \beta, \alpha + \beta \leq 1$ are hyperparameters which control relative strength of uncertainty, diversity and representativeness terms. We normalize the three scores before combining them through Eq.~\ref{eq:submod}.\\
\textbf{Lemma 1:} The set function $f(S)$ defined by Eq.~\ref{eq:submod} is submodular. \\
\textbf{Lemma 2:} The set function $f(S)$ defined by Eq.~\ref{eq:submod} is a non decreasing, monotone function.%

We provide proof of the above lemmas in Sec. 2 of supplementary material. Overview of the overall sampling approach is present in Fig. \ref{fig:sampling-al}.

\subsubsection{Submodular Optimization}
\label{sec:submod}
As we have shown in the previous section that the set function $f(S)$ is submodular, we aim to select the set $S$ satisfying the following objective:
\begin{equation}
    \label{eq:submod_obj}
    \underset{S: |S| = B}{max} f(S) 
\end{equation}
For obtaining the set of samples $S$ to be selected, we use the greedy optimization procedure. We start with an empty set $S$ and add each item iteratively. For selecting each of the sample ($x_i$) in the unlabeled set, we calculate the gain of each the sample $f(S \cup \{x_i\}) - f(S)$. The sample with the highest gain is then added to set $S$. The above iterations are done till we have exhausted our labeling budget $B$. %
The performance guarantee of the greedy algorithm is given by the following result:
\\
\textbf{Theorem 1}: Let $S^*$ be the optimal set that maximizes the objective in Eq. \ref{eq:submod_obj} then the solution $S$ found by the greedy algorithm has the following guarantee (See Supp. Sec. 2):
\begin{equation}
    f(S) \geq \left(1 - \frac{1}{e}\right)f(S^*)
\end{equation}

\textbf{Insight for Diversity Component:} The optimization algorithm with $\alpha = 0$ and 
$\beta=1$ degenerates to greedy version of diversity based Core-Set~\cite{sener2016learning} (i.e., $K$-Center Greedy) sampling. Diversity functions based on similar ideas have also been explored for different applications in \cite{joseph2019submodular, chakraborty2014adaptive}. Further details are provided in the Sec. 3 of supplementary material.

\subsection{Virtual Adversarial Active Domain Adaptation}
\label{section:VAADA}
Discriminator-alignment based Unsupervised DA methods fail to effectively utilize the additional labeled data present in target domain \cite{saito2018maximum}. This creates a need for modifications to existing methods which enable them to effectively use the additional labeled target data, and improve generalization on target data. In this work we introduce VAADA (Virtual Adversarial Active Domain Adaptation) which enhances VADA through modification which allow it to effectively use the labeled target data.

We have given our subset selection procedure to select samples to label (i.e., $\mathcal{D}_t$) in Algo. \ref{alg:main_algo} and in Fig. \ref{fig:al}. 
For aligning the features of labeled ($\mathcal{D}_s \cup \mathcal{D}_t$) with $\mathcal{D}_u$, we make use of
domain adversarial training (DANN) loss functions given below:
\vspace{-2mm}
\begin{equation}
    L_{y}(\theta; \mathcal{D}_s, \mathcal{D}_t) = \mathbb{E}_{(x,y) \sim (\mathcal{D}_s \cup \mathcal{D}_t)}[y^T\; \ln h_{\theta}(x)]
\end{equation}
\vspace{-6mm}
\begin{equation}
\begin{split}
        L_{d}(\theta; \mathcal{D}_{s}, \mathcal{D}_t, \mathcal{D}_u) = \underset{D_{\phi}}{sup} \; \mathbb{E}_{x \sim \mathcal{D}_s \cup \mathcal{D}_t}[\ln D_{\phi}(f_{\theta}(x))] \\ +\  \mathbb{E}_{x \sim \mathcal{D}_u} [\ln(1 - D_{\phi}(f_{\theta}(x)))]
\end{split}
\end{equation}
As our sampling technique is based on cluster assumption, for enforcing it we add the Conditional Entropy Loss defined in Eq.~\ref{eq:Conditional Entropy}. Additionally, for enforcing Lipschitz continuity by Virtual Adversarial Training, we use the loss defined in Eq.~\ref{eq: Vat}. 
The final loss is obtained as:
\begin{equation}
\label{eq:final-loss}
\begin{split}
    L(\theta ; \mathcal{D}_{s}, \mathcal{D}_t, \mathcal{D}_u) = L_y(\theta;{\mathcal{D}_s , \mathcal{D}_t})\ + \\ \lambda_dL_d(\theta; \mathcal{D}_s, \mathcal{D}_t, \mathcal{D}_u)\ + \lambda_sL_{v}(\theta; \mathcal{D}_s \cup \mathcal{D}_t)\ +\\ \lambda_t(L_v(\theta; \mathcal{D}_u) + L_c(\theta; \mathcal{D}_u))
\end{split}
\end{equation}

The $\lambda$-values used are the \textit{same for all our experiments} and are mentioned in the Sec. 5 of supplementary material.

\textbf{Differences between VADA and VAADA}:
We make certain important changes to VADA listed below, which enables VADA \cite{shu2018dirt}  to effectively utilize the additional supervision of labeled target data and for VAADA procedure:
\\ \textbf{1) High Learning Rate for All Layers}: In VAADA, we use the same learning rate for all layers. In a plethora of DA methods \cite{saito2018maximum, long2018conditional} a lower learning rate for initial layers is used to achieve the best performance. We find that although this practice helps for Unsupervised DA it hurts the Active DA performance (experimentally shown in Sec. 5 of supplementary material).
\\  \textbf{2) Using Gradient Clipping in place of Exponential Moving Average (EMA)}: We use gradient clipping for all network weights to stabilize training whereas VADA uses EMA for the same. We find that clipping makes training of VAADA stable in comparison to VADA and achieves a significant performance increase over VADA.

We find VAADA is able to work robustly across diverse datasets. It has been shown in \cite{saito2018maximum} that VADA, when used out of the box, is unable to get gains in performance when used in setting where target labels are also available for training. This also agrees with our observation that VAADA significantly outperforms VADA in Active DA scenario's (demonstrated in Fig. \ref{fig:training-ablation}, with additional analysis in Sec. 5 of supplementary material).

\begin{algorithm}
\label{alg:main_algo}
\caption{S$^3$VAADA: Submodular Subset Selection for Virtual Adversarial Active Domain Adaptation}
\begin{algorithmic}[1]
\REQUIRE Labeled source $\mathcal{D}_s$; Unlabeled target  $\mathcal{D}_u$; Labeled target $\mathcal{D}_t$; Budget per cycle $B$; Cycles $C$; Model with parameters $\theta$; Parameters $\alpha$, $\beta$
\ENSURE Updated model parameters with improved generalization ability on target domain
\STATE {Train the model according to Eq.~\ref{eq:final-loss}}
\FOR{cycle $\gets$ 1 to $C$}
\STATE {$S$ $\leftarrow$ $\emptyset$}
\FOR{iter $\gets$ 1 to $B$}
\STATE{$x^* = \underset{x \in \mathcal{D}_u \setminus S}{argmax} \; f(S \cup \{x\}) - f(S)$}
\STATE{$S$ $\leftarrow$ $S$ $\cup$ \{$x^*$\}}
\ENDFOR
\STATE{Get ground truth labels $l_{S}$ for samples in $S$ from oracle}
\STATE{$\mathcal{D}_t \leftarrow \mathcal{D}_t \cup (S, l_{S})$}
\STATE{$\mathcal{D}_u \leftarrow \mathcal{D}_u \setminus S$}
\STATE{Train the model according to Eq.~\ref{eq:final-loss}}
\ENDFOR
\end{algorithmic}
\end{algorithm}

\section{Experiments}
\label{sec:experiments}
\subsection{Datasets}
We perform experiments across multiple source and target domain pairs belonging to 3 diverse datasets, namely Office-31 \cite{saenko2010adapting}, Office-Home \cite{venkateswara2017Deep}, and VisDA-18 \cite{8575439}. We have specifically not chosen any DA task using real world domain as in those cases the performance maybe higher due to ImageNet initialization not due to adaptation techniques. In \textbf{Office-31} dataset, we evaluate the performance of various sampling techniques on DSLR $\rightarrow$ Amazon and Webcam $\rightarrow$ Amazon, having 31 classes. 
The \textbf{Office-Home} consists of 65 classes and has 4 different domains belonging to Art, Clipart, Product and Real World. We perform the active domain adaptation on Art $\rightarrow$ Clipart, Art $\rightarrow$ Product and Product $\rightarrow$ Clipart. 
\textbf{VisDA-18} image classification dataset consists of two domains (synthetic and real) with 12 classes in each domain. %

\subsection{Experimental Setup}
Following the common practice in AL literature, we first split the target dataset into train set and validation set with 80\% data being used for training and 20\% for validation. 
In all the experiments, we set budget size $B$ as 2\% of the number of images in the target training set, and we perform five cycles ($C$ = 5) of sampling in total. At the end of all cycles, 10\% of the labeled target data will be used for training the model. This setting is chosen considering the practicality of having a small budget of labeling in the target domain and having access to unlabeled data from the target domain. We use ResNet-50 \cite{He_2016_CVPR} as the feature extractor $g_{\theta}$, which is initialized with weights pretrained on ImageNet.
We use SGD Optimizer with a learning rate of 0.01 and momentum (0.9) for VAADA training. 
For the DANN experiments, we follow the same architecture and training procedure as described in \cite{long2018conditional}.
In all experiments, we set $\alpha$ as 0.5 and $\beta$ as 0.3.
We use PyTorch \cite{NEURIPS2019_9015} with automatic mixed-precision training for our implementation. 
Further experimental details are described in the Sec. 6 of  supplementary material.
We report the mean and standard error of the accuracy of the 3 runs with different random seeds.

In AADA \cite{Su_2020_WACV} implementation, the authors have used a different architecture and learning schedule for DANN  which makes comparison of increase in performance due to Active DA, over unsupervised DA intricate. In contrast we use ResNet-50 architecture and learning rate schedule of DANN used in many works \cite{long2018conditional, chen2020adversariallearned}. We first train DANN to reach optimal performance on Unsupervised Domain Adaptation and then start the active DA process. This is done as the practical utility of Active DA is the performance increase over Unsupervised DA. 
\subsection{Baselines}
\begin{figure}[!t]
  \centering
  \includegraphics[width=\linewidth]{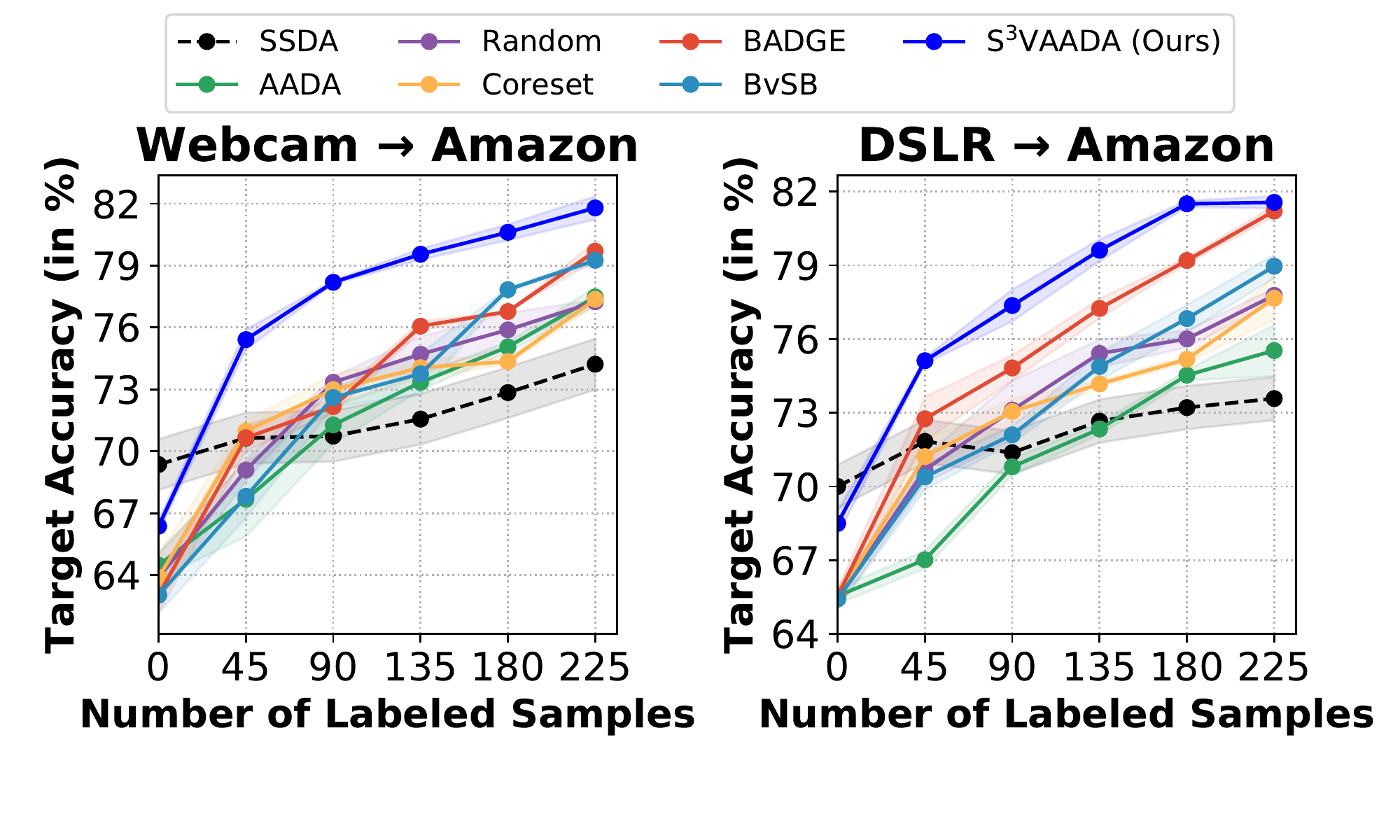}
  \vspace{-10mm}
  \caption{Active DA target accuracy on two adaptation tasks from Office-31 dataset. S$^3$VAADA consistently outperforms BADGE~\cite{Ash2020Deep}, AADA~\cite{Su_2020_WACV} and SSDA (MME$^*$~\cite{saito2019semi}) techniques.}
  \label{fig:office-31-results}
\end{figure}
It has been shown by Su et al. \cite{Su_2020_WACV} that for active DA performing adversarial adaptation through DANN, with adding newly labeled target data into source pool works better than fine-tuning. Hence, we use DANN for all the AL baselines described below: 

\begin{figure*} [!t]
  \centering
  \includegraphics[width=\linewidth]{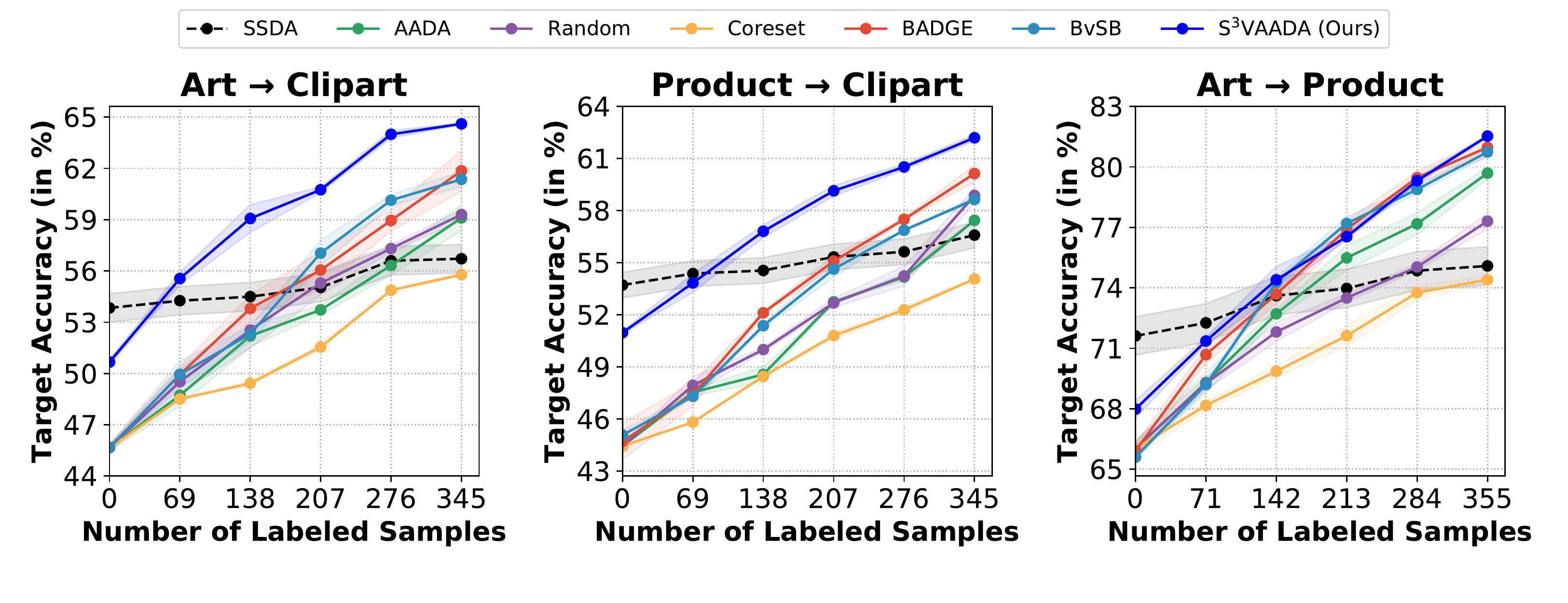}
  \vspace{-10mm}
  \caption{Active DA performance on three different Office-Home domain shifts. We see a significant improvement through S$^3$VAADA in two difficult adaptation tasks of  Art $\rightarrow$ Clipart (left) and Product $\rightarrow$ Clipart (middle) .}
  \label{fig:office-home-results}
\end{figure*}

\begin{enumerate}[noitemsep]
\itemsep0em
    \item \textbf{AADA} (Importance weighted sampling) \cite{Su_2020_WACV}: This state-of-the-art active DA method incorporates uncertainty by calculating entropy and incorporates diversity by using the output of the discriminator.
    \item \textbf{BvSB} (Best vs Second Best a.k.a. margin) \cite{5206627}: It uses the difference between the probabilities of the highest and second-highest class prediction as the the metric of uncertainty, on which low value indicates high uncertainty.
    \item \textbf{BADGE} \cite{Ash2020Deep}: BADGE incorporates uncertainty and diversity by using the gradient embedding on which k-MEANS++ \cite{vassilvitskii2006k} algorithm is used to select diverse samples. BADGE method is currently one of the state-of-the-art methods for AL.
    \item \textbf{$K$-Center (Core-Set)} \cite{sener2018active}: Core-Set selects samples such that the area of coverage is maximized. We use greedy version of Core-Set on the feature space ($g_{\theta}$). It is a diversity-based sampling technique.
    \item \textbf{Random}: Samples are selected randomly from the pool of unlabeled target data.

\end{enumerate}
\vspace{-2mm}
\textbf{Semi-Supervised DA:} We compare our method against recent method of MME$^*$~\cite{saito2019semi} with ResNet-50 backbone on Office datasets, using the author's implementation\footnote{https://github.com/VisionLearningGroup/SSDA\_MME}. In each cycle target samples are randomly selected, labeled and provided to the MME$^*$ method for DA.

\subsection{Results}

\begin{figure}[!t]
  \centering
  \includegraphics[width=0.8\linewidth,height=5cm]{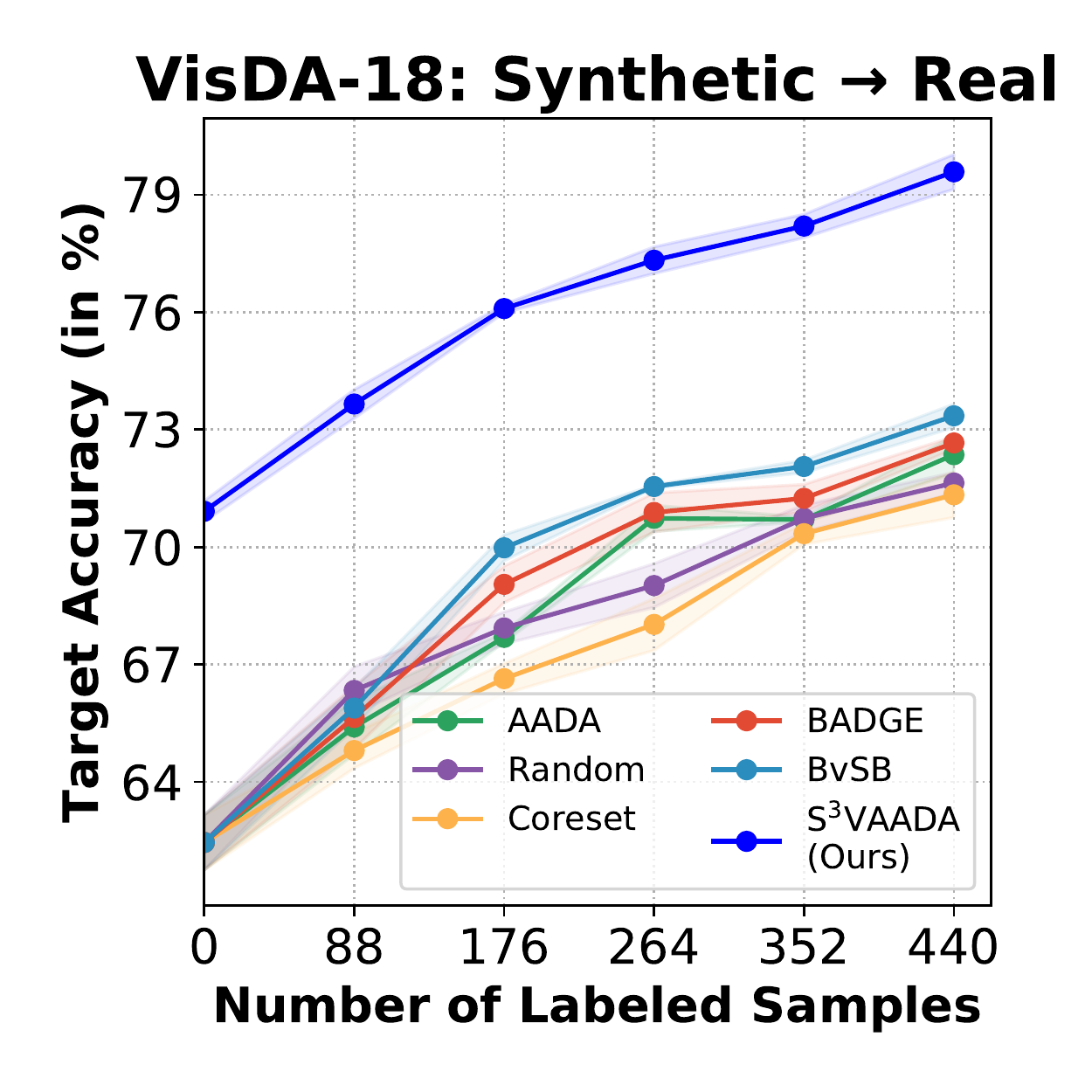}
  \vspace{-3mm}
  \caption{Active DA Results on VisDA-18 dataset.}
  \label{fig:visda-results}
\end{figure}

\begin{figure*}[!t]
\begin{minipage}{0.75\linewidth}
    \centering
 \subfigure[Uncertainty (\textbf{U})]{\includegraphics[width=0.3\linewidth,height=4cm]{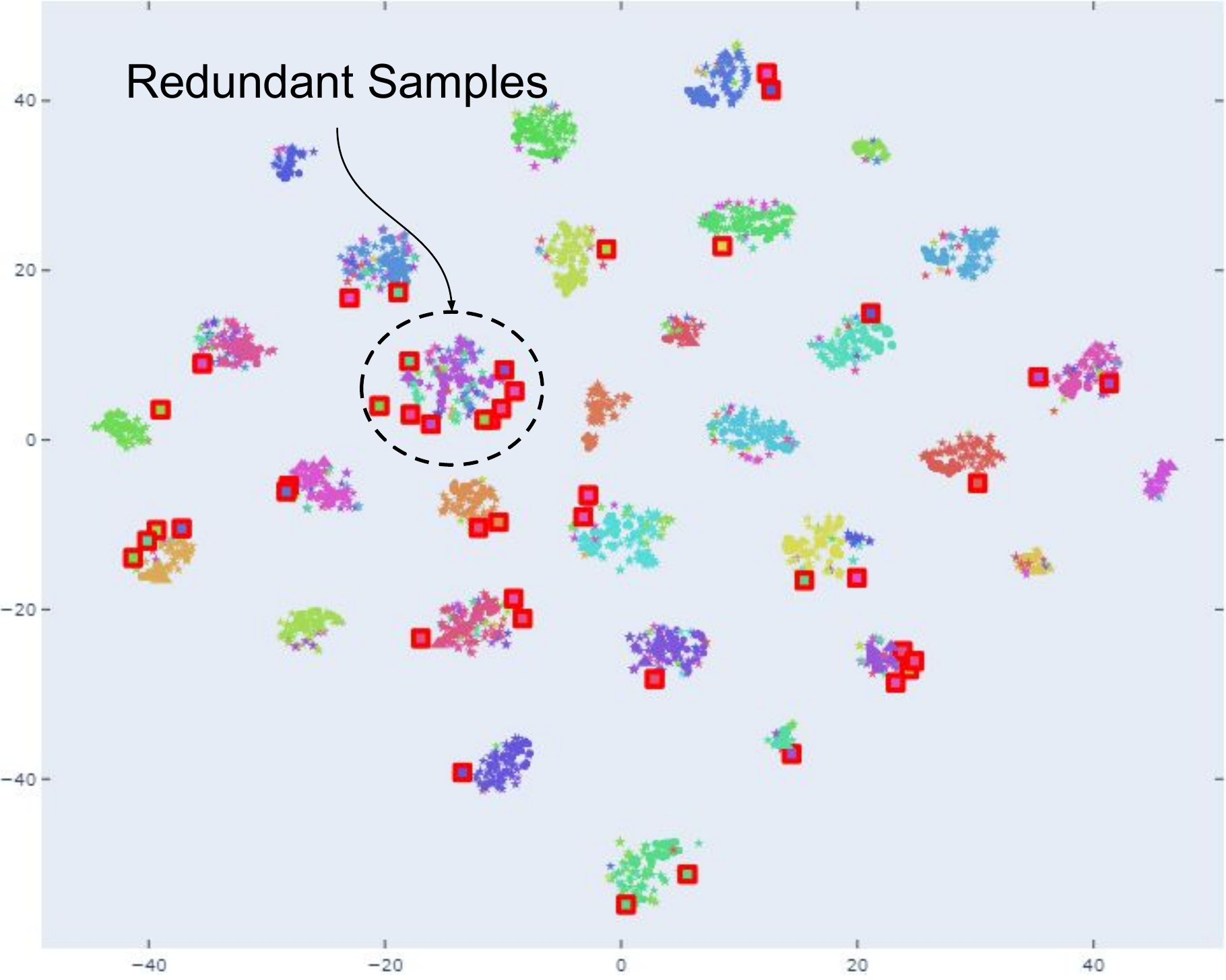}}
  \subfigure[Diversity (\textbf{D})]{\includegraphics[width=0.3\linewidth,height=4cm]{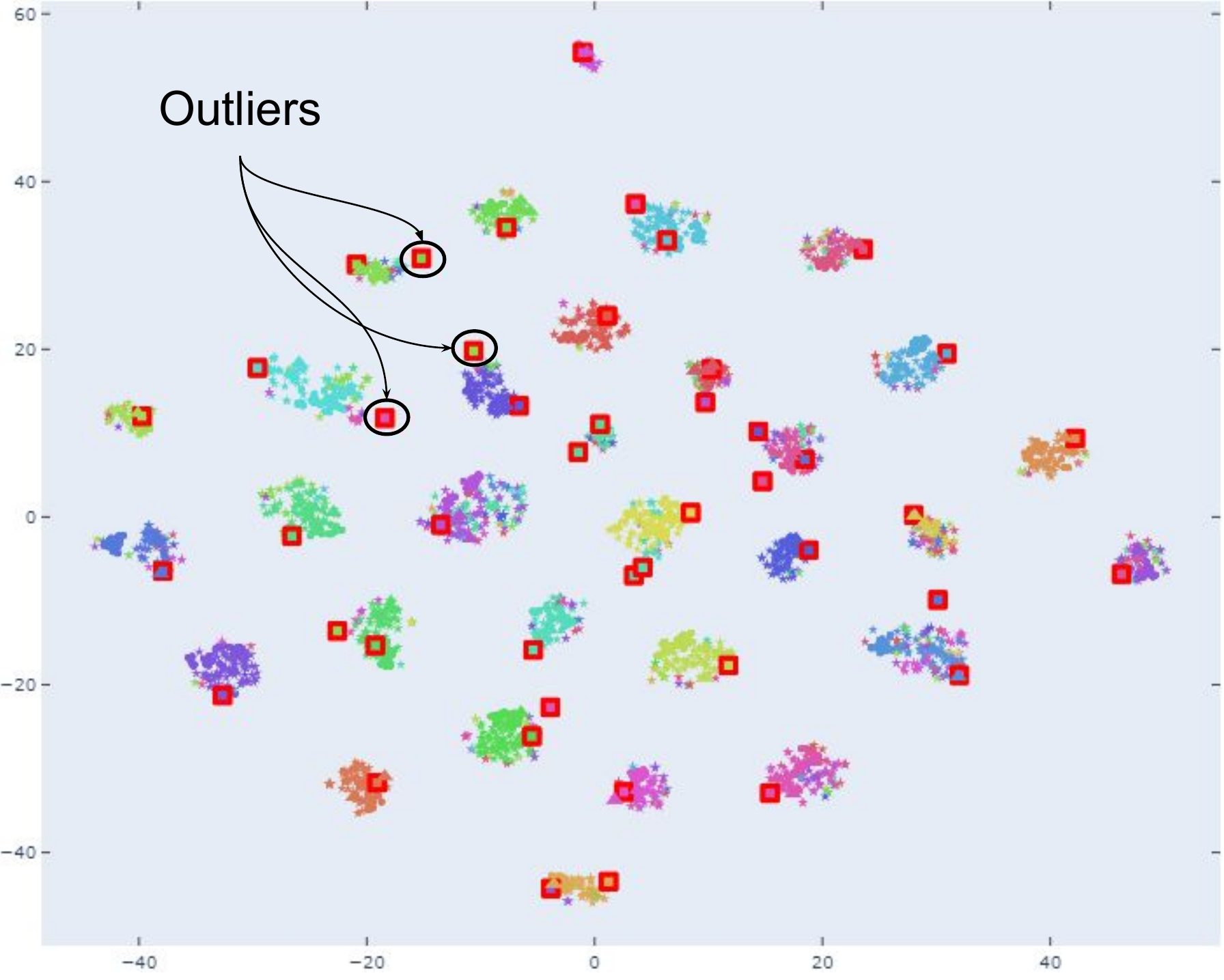}}
  \subfigure[Representativeness (\textbf{R})]{\includegraphics[width=0.3\linewidth,height=4cm]{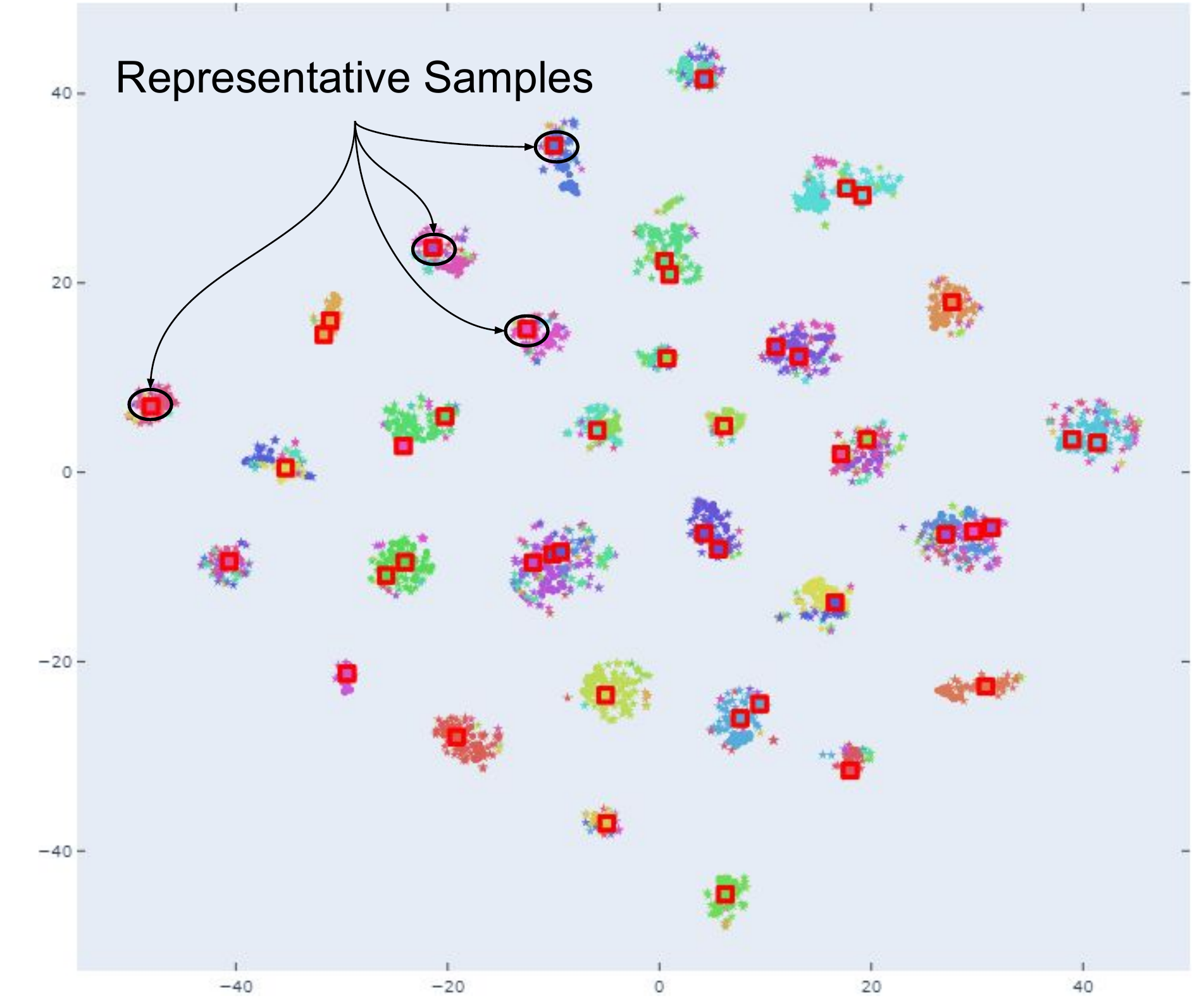}}
  \caption{Feature space visualization using t-SNE with selected samples in Red. Using \textbf{Uncertainty} leads to \textit{redundant} samples from same cluster, whereas using \textbf{Diversity} leads to only diverse boundary samples being selected which maybe \textit{outliers}. 
  Sampling using \textbf{Representativeness} prefers samples near the cluster center, hence we use a combination of these complementary criterion as our criterion.}
  \label{fig:tsne-plot}
\end{minipage}%
\hspace{1mm}
\begin{minipage}{0.24\textwidth}
\centering
      {\includegraphics[width=\linewidth]{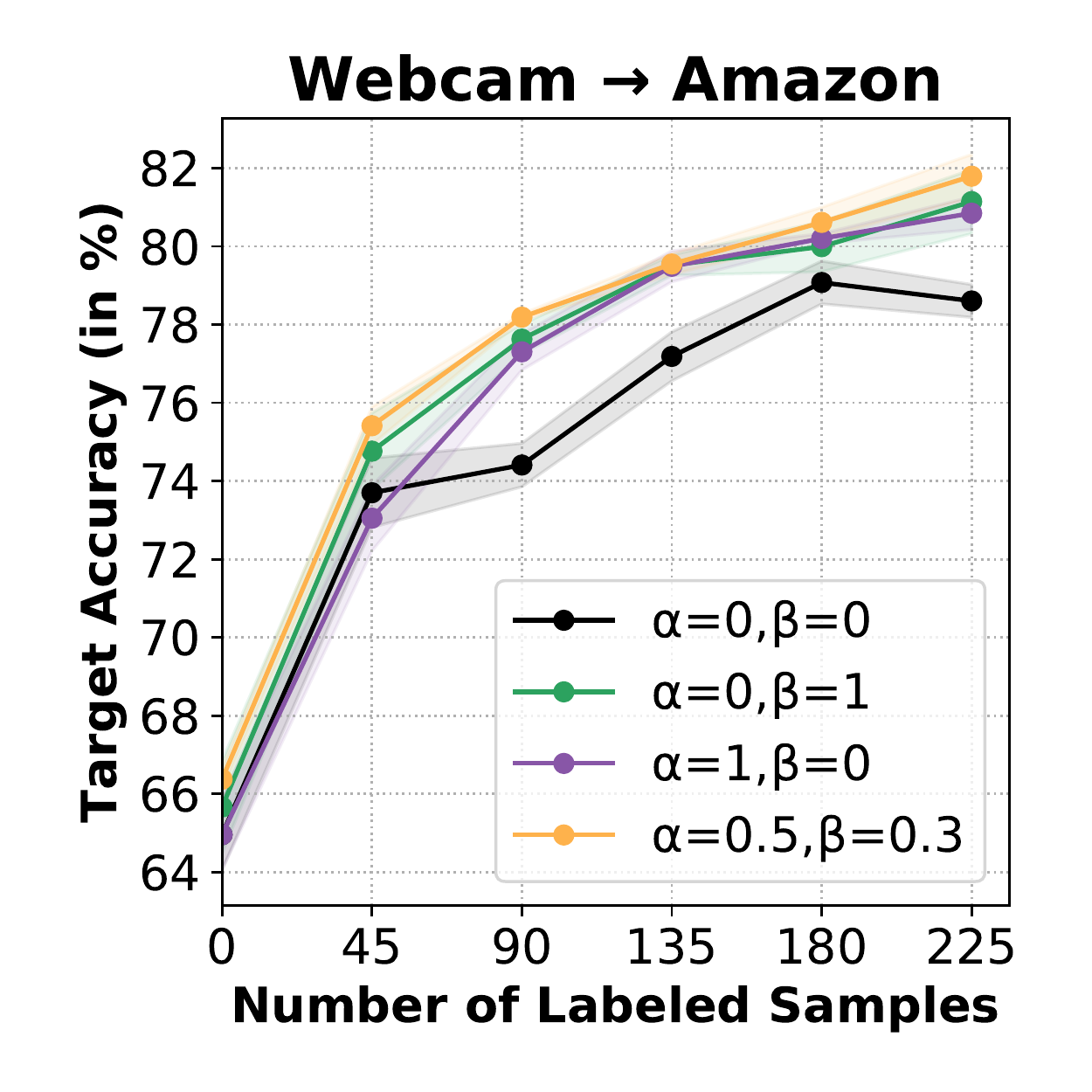}}
  \caption{Trade off between Uncertainty, Diversity and Representativeness (i.e., Parameter sensitivity to $\alpha, \beta$).}
  \label{fig:alpha-beta}
\end{minipage}

\end{figure*}

\begin{figure}[!t]
  \centering
  \includegraphics[width=\linewidth]{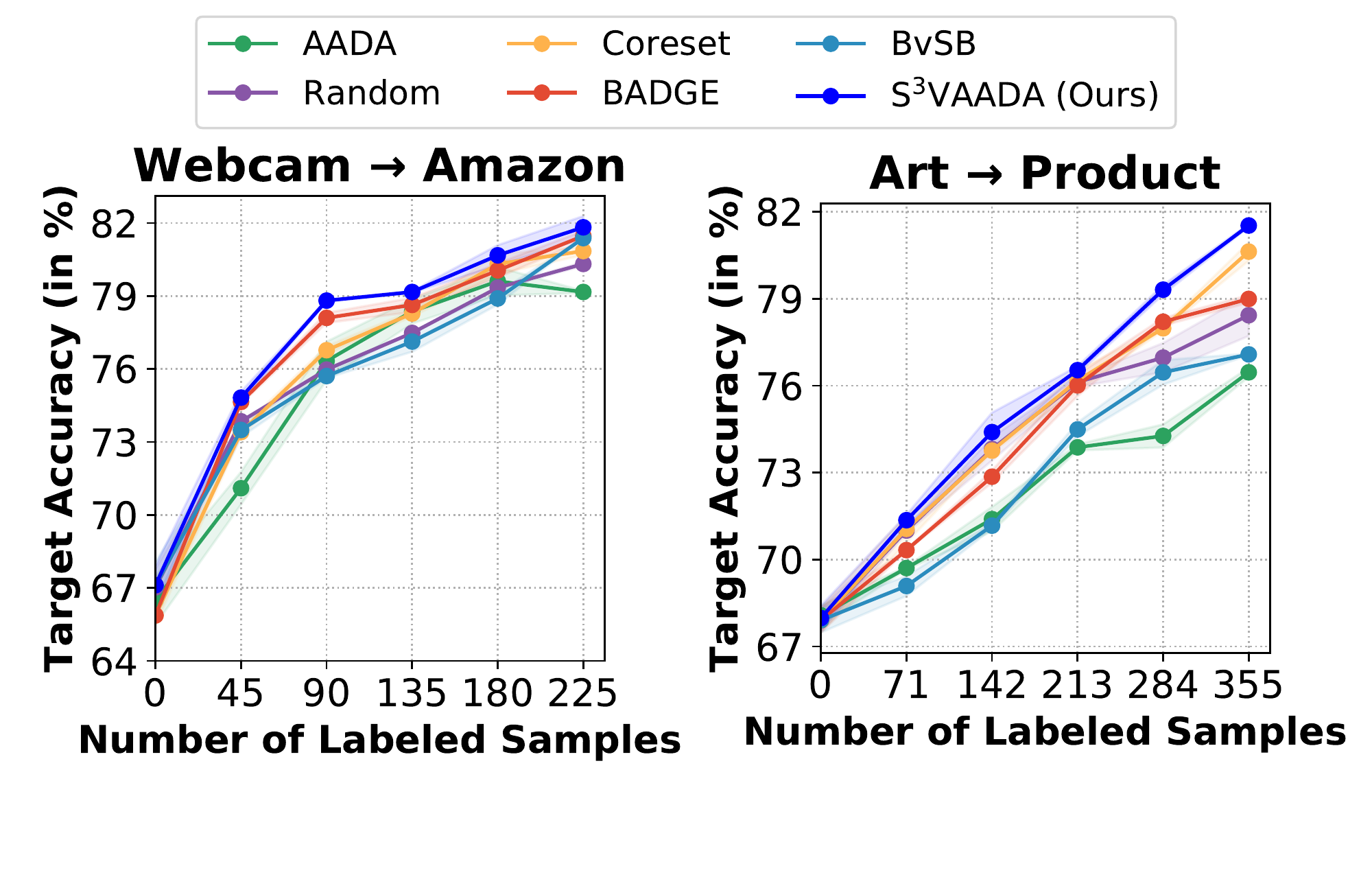}
  \vspace{-10mm}
  \caption{Ablation on sampling methods on different domain shifts. In both cases, we train the sampling techniques via VAADA. Our method (S$^3$VAADA) consistently outperforms all other sampling methods.}
  \label{fig:sampling-ablation}
\end{figure}

\label{sec:results}
Fig. \ref{fig:office-31-results} shows the results on Office-31 dataset, S$^3$VAADA outperforms all other techniques. On Webcam $\rightarrow$ Amazon shift, it shows significant improvement of 9\% in the target accuracy with just 45 labeled samples. S$^3$VAADA gets 81.8\% accuracy in the last cycle which is around 15\% more than the unsupervised DA performance, by using just 10\% of the labeled data. On DSLR $\rightarrow$ Amazon shift, VAADA follows a similar trend and performs better than all other sampling techniques. On Office-Home dataset on the harder domain shifts of Art $\rightarrow$ Clipart and Product $\rightarrow$ Clipart, S$^3$VAADA produces a significant increase of 3\%-5$\%$ and 2\%-6\% respectively across cycles, in comparison to other methods (Fig. \ref{fig:office-home-results}). On the easier Art $\rightarrow$ Product shift, our results are comparable to other methods.

\textbf{Large Datasets:} On the VisDA-18 dataset, where the AADA method is shown to be ineffective \cite{Su_2020_WACV} due to a severe domain shift. Our method (Fig. \ref{fig:visda-results}) is able to achieve significant increase of around 7\% averaged across all cycles, even in this challenging scenario. For demonstrating the scalability of our method to DomainNet~\cite{peng2019moment}, we also provide the results of one adaptation scenario in Sec. 10 of supplementary material. 

\textbf{Semi-Supervised DA:} From Figs. \ref{fig:office-31-results} and \ref{fig:office-home-results} it is observed that performance of MME$^*$ saturates as more labeled data is added, in contrast, S$^3$VAADA continues to improve as more target labeled data is added. 

\section{Analysis of S$^3$VAADA}
\textbf{Visualization:} Fig. \ref{fig:tsne-plot} shows the analysis of samples selected for uncertainty \textbf{U}, diversity \textbf{D} and representativeness \textbf{R} criterion, which depicts the \textit{complementary} preferences of the three criterion. \\
\textbf{Sensitivity to $\alpha$ and $\beta$ }: Fig. \ref{fig:alpha-beta} shows experiments for probing the effectiveness of each component (i.e., uncertainty, diversity and representativeness) in the information criteria.
We find that just using Uncertainty ($\alpha = 1$) and Diversity ($\beta = 1$) provide reasonable results when used individually. However, the individual performance remain sub-par with the hybrid combination (i.e., $\alpha = 0.5$, $\beta=0.3$). We use value of $\alpha=0.5$ and $\beta=0.3$ across all our experiments, hence our sampling does not require parameter-tuning specific to each dataset.

\textbf{Comparison of Sampling Methods:}
For comparing the different sampling procedures we fix the adaptation technique to VAADA and use different sampling techniques. Fig. \ref{fig:sampling-ablation} shows that our sampling method outperforms others in both cases. In general we find that hybrid approaches i.e., Ours and BADGE perform \textit{robustly} across domain shifts.

\begin{figure}[!t]
  \centering
  \includegraphics[width=\linewidth]{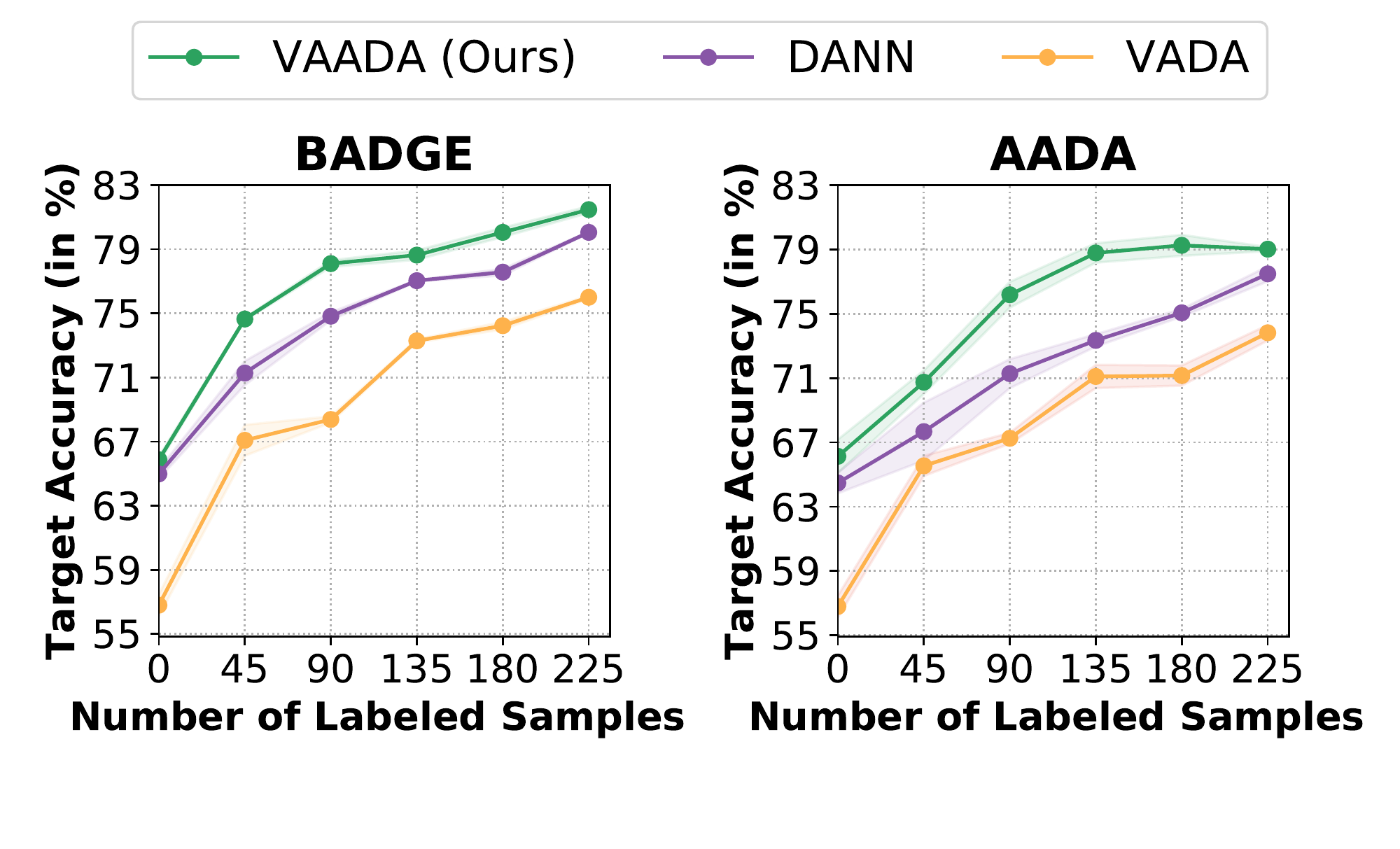}
  \vspace{-11mm}
  \caption{Comparison of different DA methods for Active DA on Webcam $\rightarrow$ Amazon.}
  \label{fig:training-ablation}
\end{figure}

\textbf{Comparison of VAADA:} Fig. \ref{fig:training-ablation} shows performance of VAADA, DANN and VADA when used as adaptation procedure for two sampling techniques. We find that a significant improvement occurs for all the sampling techniques in each cycle for VAADA comparison to DANN and VADA. The $\geq 5\%$ improvement in each cycle,  shows the importance of proposed improvements in VAADA over VADA.

We provide additional analysis on \textit{convergence, budget size and hyper parameters} in the Sec. 4 of supplementary material. Across our analysis we find that S$^3$VAADA works robustly in different DA scenario.

\section{Conclusion}
We formulate the sample selection in Active DA as optimal informative subset selection problem, for which we propose a novel submodular information criteria. The information criteria takes into account the uncertainty, diversity and representativeness of the subset. The most informative subset obtained through submodular optimization is then labeled and used by the proposed adaptation procedure VAADA. We find that the optimization changes introduced for VAADA significantly improve Active DA performance across all sampling schemes. The above combination of sampling and adaptation procedure constitutes S$^3$VAADA, which consistently provides improved results over existing methods of Semi-Supervised DA and Active DA.  \\
\textbf{Acknowledgements:} This work was supported in part by PMRF Fellowship (Harsh), SERB (Project:STR/2020/000128) and UAY (Project:UAY, IISC\_010) MHRD, Govt. of India.

\clearpage

\appendix

\renewcommand \thepart{}
\renewcommand \partname{}

\doparttoc 
\faketableofcontents 
\appendix
\addcontentsline{toc}{section}{Appendix} 
\part{Supplementary Material} 
\parttoc 

\section{t-SNE Analysis for AADA}
\label{aada}
\begin{figure}[h]
    \centering
    \includegraphics[width=0.49\textwidth]{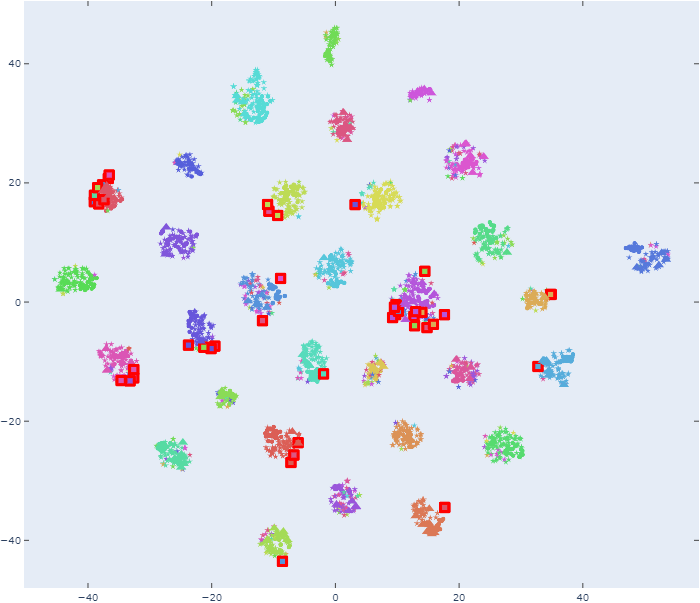}
    \caption{t-SNE analysis of AADA sampling. The selected samples are represented by the red boxes. We see clusters of samples being selected which depict \textit{redundancy} in selection. }
    \label{fig:tsne-aada}
\end{figure}

We give experimental evidence of the redundancy issue present in the AADA sampling. We perform the training with VAADA training method with the implementation details present in Sec. \ref{experimental_details} on Webcam $\rightarrow$ Amazon. Fig. \ref{fig:tsne-aada} shows the selected samples in the intermediate cycle, which clearly depicts clusters of the samples selected. The existence of clusters confirms the presence of \textit{redundancy} in selection.

\section{Proofs}
\label{proof}
\subsection{Lemma 1}
We present proof for lemma 1 which is stated in Section {\color{red} 4.1.4} of the main paper.

\textbf{Lemma 1} The set function $f(S)$ defined by equation below is submodular.  \\
\begin{multline*}
        f(S \cup \{x_i\}) - f(S) = \alpha VAP(x_i) + \beta d(S,x_i) \\ +
     (1 - \alpha - \beta) R(S, x_i)
\end{multline*}
    
We first prove that all the three individual components in the above expression are submodular and then prove that the convex combination of the three terms is submodular.

\textbf{Submodularity of the VAP Score $VAP(x_i)$:}
The gain value of the VAP score is given as the following below:
$$
f(S \cup \{x_i\}) - f(S) = VAP(x_i)
$$
We give below the proof for the submodularity which is based on the \textit{diminishing returns} property as stated in Sec. {\color{red} 3.1}, in the main text.
\begin{proof}
  For two sets $S_1, S_2$ such that $S_1 \subseteq S_2$ and $x_i \in \Omega \backslash S_2$, if the function is submodular it should satisfy the following property in Sec {\color{red} 3.1}.
      \begin{align*}
        f(S_1 \cup \{x_i\}) - f(S_1) &\geq f(S_2 \cup \{x_i\}) - f(S_2) \\
        VAP(x_i) &\geq  VAP(x_i)
    \end{align*}
    As the left hand side is equal to right hand side, the inequality is satisfied, hence the VAP score function is submodular.
\end{proof}
\textbf{Submodularity of Diversity Score $d(S, x_i)$:}
The gain in value for the diversity function is given as:
$$
f(S \cup \{x_i\}) - f(S) = \min_{x \in S} D(x, x_i)
$$
We provide the proof that the above gain function corresponds to a submodular function $f(S)$ below:
\begin{proof}
For two sets $S_1, S_2$ such that $S_1 \subseteq S_2$ and $x_i \in \Omega \backslash S_2$, if the function is submodular it should satisfy the following property in Sec {\color{red} 3.1}:
    \begin{align*}
        f(S_1 \cup \{x_i\}) - f(S_1) &\geq f(S_2 \cup \{x_i\}) - f(S_2) \\
        min_{x \in S_1} D(x, x_i) &\geq  min_{x \in S_2} D(x, x_i)
    \end{align*}
$D(x, x_i) \geq 0$ for every $x$ and $x_i$ as it is a divergence function. As $S_2$ contains more elements than $S_1$, the minimum of $D(x, x_i)$  will be less then for $S_2$ in comparison to that of $S_1$. Hence the final inequality is satisfied which shows that $f(S)$ is submodular.
\end{proof}
\textbf{Submodularity of Representativeness Score $R(S, x_i)$:}
We first prove one property which we will use for analysis of Representativeness Score. 
\begin{lemma}
\textit{The sum of two submodular set functions $f(S)$ = $f_1(S) + f_2(S)$, is submodular.}
\end{lemma}
\begin{proof}
Let A and B be any two random sets.
  \begin{align*}
    f(A) + f(B) &= f_1(A) + f_2(A) + f_1(B) + f_2(B) \\
                &\geq f_1(A \cup B) + f_2(A \cup B) + f_1(A \cap B) + \\ &\; f_2(A \cap B) \\
                &= f(A \cup B) + f(A \cap B)
  \end{align*}
  Hence the sum of the two submodular functions is also submodular. The result can be generalized to a sum of arbitrary number of submodular functions.
\end{proof}

The representativeness score can be seen as the following set function below:

\begin{align*}
 f(S)  &= \underset{x_i \in \mathcal{D}_u}{\sum} \underset{x_j \in S}{\max}\ s_{ij}  
\end{align*}
We calculate the gain for each sample through this function which is equal to $R(S, x_i)$:
\begin{align*}
    f(S \cup \{x_i\}) - f(S) &=  \underset{x_k \in \mathcal{D}_u}{\sum} \underset{x_j \in S \cup \{x_i\}}{\max} s_{kj} - \underset{x_k \in \mathcal{D}_u}{\sum} \underset{x_j \in S}{\max }\ s_{kj} \\
    R(S, x_i) &= \underset{x_k \in \mathcal{D}_u}{\sum} \max(s_{ik} - \underset{x_j \in S}{\max}\ s_{kj}, 0)
\end{align*}
\begin{lemma}
\textit{ The set function defined below is submodular:}
$$f(S) = \underset{x_i \in \mathcal{D}_u}{\sum} \underset{x_j \in S}{\max}\ s_{ij}$$
\end{lemma}
\begin{proof}
  We first show that the function $f_i(S) = \underset{x_j \in S}{\max}\ s_{ij}$ is submodular. We first use the property, $f(A) + f(B) \geq f(A \cup B) + f(A \cap B)$ where $A, B$ are two sets, sufficient to show that $f(S)$ is submodular: 
\begin{align}
    f_i(A) + f_i(B) &\geq f_i(A \cup B) + f_i(A \cap B) \\
    \underset{x_j \in A}{\max}\ s_{ij} +  \underset{x_j \in B}{\max}\ s_{ij} &\geq \underset{x_j \in A \cup B}{\max} s_{ij} + \underset{x_j \in A \cap B}{\max} s_{ij} 
 \end{align}
 which follows due to the following:
 $$
 \max(\underset{x_j \in A}{\max}\ s_{ij}, \underset{x_j \in B}{\max}\ s_{ij}) =  \max_{x_j \in A \cup B} s_{ij}
 $$
 and 
 $$
 \min(\underset{x_j \in A}{\max}\ s_{ij}, \underset{x_j \in B}{\max}\ s_{ij}) \geq \max_{x_j \in A \cap B} s_{ij}
 $$
As $f_i(S)$ is submodular, the $f(S)$ can be seen as:
$$
f(S) = \sum_{x_i \in \mathcal{D}_u} f_i(S)
$$
which is submodular according to the property that sum of submodular functions is also submodular proved above.
\end{proof}
 \textbf{Combining the Submodular Functions:} We use the property that a convex combination of the submodular functions is also submodular.
Hence our sampling function which is the convex combination given by:
\begin{multline*}
    f(S \cup \{x_i\}) - f(S) = \alpha VAP(x_i) + \beta d(S, x_i)  \\ + (1 - \alpha - \beta) R(S, x_i)
\end{multline*}

Also follows the property of submodularity.

\subsection{Lemma 2}
Here we present proof of lemma 2 stated in Sec. {\color{red} 4.1.4} of main paper. \\
\textbf{Lemma 2} The set function f(S) defined by equation below is a non-decreasing, monotone function:
    \begin{align*}
    f(S \cup \{x_i\}) - f(S) = \alpha VAP(x_i) + \beta d(S,x_i) \\ +
     (1 - \alpha - \beta) R(S, x_i)
    \end{align*}
    
\begin{proof}
  For the function to be non-decreasing monotone for every set $S$ the addition of a new element should increase value of $f(S)$. The gain function for $f(S)$ is given below:
 \begin{align*}
     &f(S \cup \{ x_i \}) - f(S) \geq 0 \\
     &\alpha VAP(x_i) + \beta d(S,x_i) +
     (1 - \alpha - \beta) R(S, x_i) \geq 0
 \end{align*}
 As the $VAP(x_i)$ and $d(S,x_i)$ are KL-Divergence terms, they have value $\geq$ 0. The third term $R(S, x_i) = \underset{x_k \in \mathcal{D}_u}{\sum} \max(s_{ik} - \underset{x_j \in S}{\max}\ s_{kj}, 0)$ is also $\geq$ 0.
 As $0 \leq \alpha, \beta, \alpha + \beta \leq 1$, the value of gain is positive, this shows that the function $f(S)$ is a non-decreasing monotone.
 
\end{proof}

\subsection{Theorem 1}
\textbf{Theorem 1}: Let $S^*$ be the optimal set that maximizes the objective in Eq. \ref{eq:submod_obj2} then the solution $S$ found by the greedy algorithm has the following approximation guarantee:
\begin{equation}
    f(S) \geq \left(1 - \frac{1}{e}\right)f(S^*)
    \label{eq:submod_obj2}
\end{equation}
\textbf{Proof: } As $f(S)$ is submodular according to Lemma 1 and is also non decreasing, monotone according to Lemma 2. Hence the approximation result directly follows from Theorem 4.3 in \cite{nemhauser1978analysis}. The approximation result shows that the algorithm is guaranteed to get at least $63\%$ of the score of the optimal function $f(S^*)$. However, in practice, this algorithm is often able to achieve $98\%$ of the optimal value in certain applications \cite{krause2009optimizing}.
As it's a worst case result in practice we get better performance than the worst case. 

\section{Insight for Diversity Score}
\label{diversity}
When the $\alpha = 0$ and $\beta=1$ the gain function $f(S \cup \{x_i\}) - f(S)$ is just $min_{x \in S} D(x,x_i)$. The greedy algorithm described for sampling in Algorithm {\color{red} 1} in main paper, leads to following objective
for selecting sample $x{^*}$.
\begin{equation*}
    x{^*} = \underset{x_i \in \mathcal{D}_u \setminus S}{argmax} \; \underset{x \in S}{\min} \; D(x,x_i)
\end{equation*}
This objective exactly resembles the $K$-Center Greedy method objective which is used by Core-Set method \cite{sener2018active} and is shown to select samples which cover the entire dataset. The $K$-Center Greedy method is very effective in practice. This connection shows that diversity component in our framework also tries to cover the dataset as done by Core-Set \cite{sener2016learning} method which is one of the very effective diversity based active learning method.

\section{Additional Analysis for S$^3$VAADA}
In this sections we provide additional experiments for analysis of the proposed S$^3$VAADA. Unless specified, we run the experiments with single random seed and report the performance. In case the performance difference is small, we provide average results of three runs with different random seeds.
\label{add_analysis}

\begin{figure}[]
  \centering
  \includegraphics[width=0.75\linewidth,height=5cm]{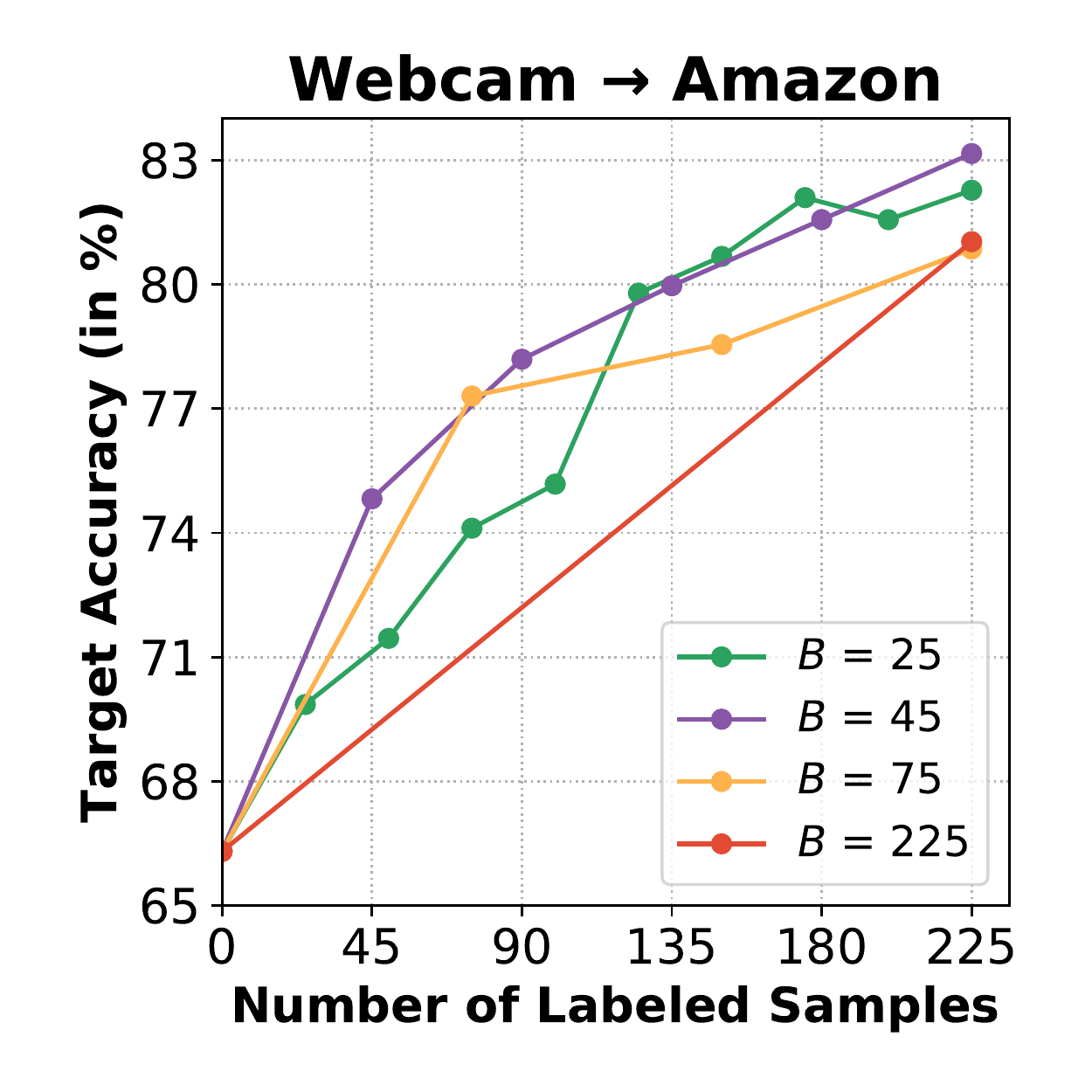}
  \caption{Analysis of S$^3$VAADA for different budget sizes on Webcam $\rightarrow$ Amazon shift of Office-31 dataset.}
  \label{fig:budget-ablation}
\end{figure}

\subsection{Budget Ablation}
Keeping in mind the practical constraint of only having a small amount of labeling budget in the target domain, we restrict ourselves to having a budget size of $2\%$ of the labeled target data. Due to different size of target data in each dataset, the sampling algorithm needs to work robustly under different budget scenario's. For further analysis, we provide results on Webcam to Amazon with different budget sizes $B$ for sampling in Fig. \ref{fig:budget-ablation}. We find that S$^3$VAADA is quite robust for budget sizes greater than 45. We find that small budget of 25 results in more stochasticity in the results.

\begin{figure}[!t]
  \centering
  \includegraphics[width=0.75\linewidth,height=5cm]{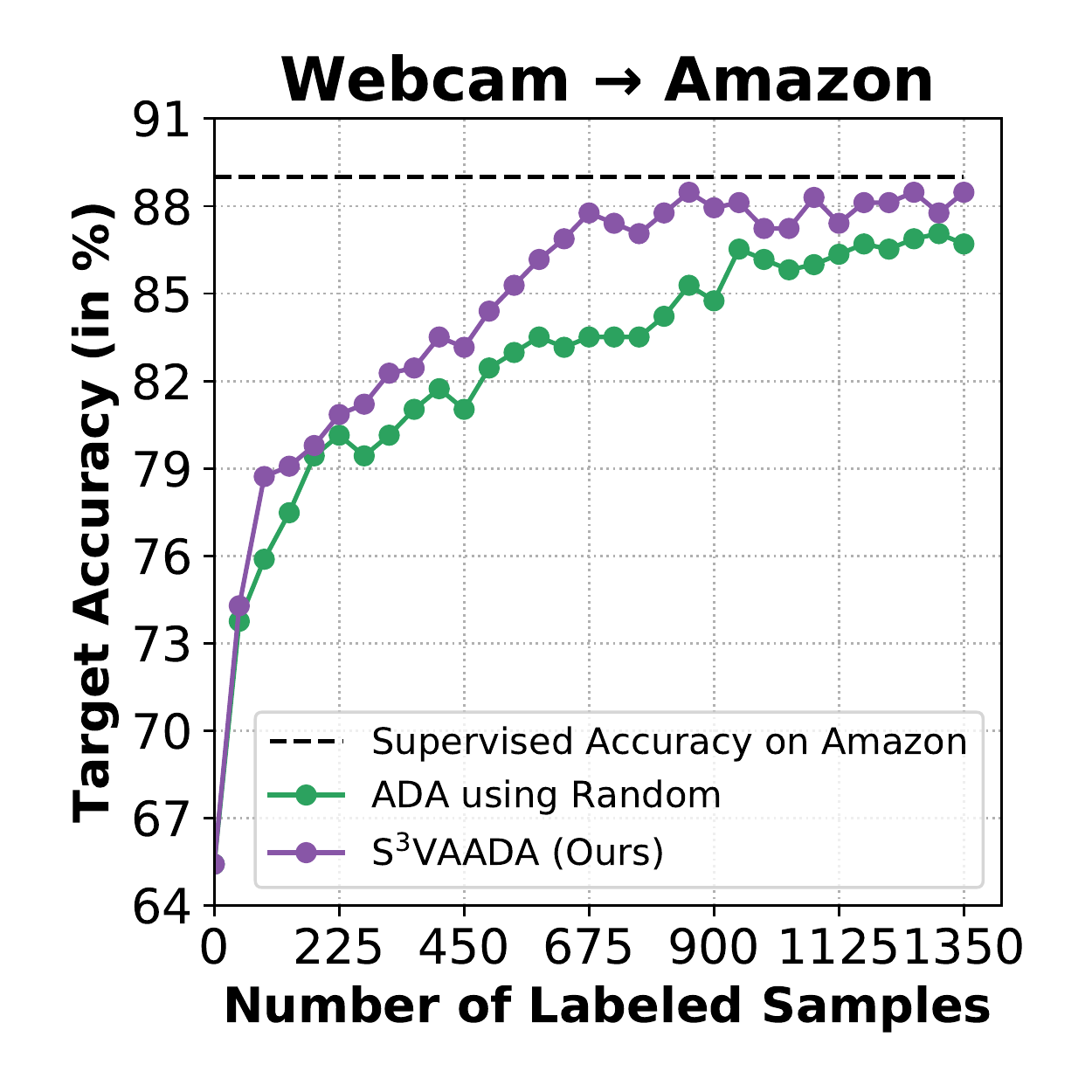}
  \caption{Active DA performance on Webcam $\rightarrow$ Amazon for 30 cycles. We find that the performance converges to supervised learning performance after around 15 cycles.}
  \label{fig:conv}
\end{figure}
\subsection{Convergence: When does the Active DA performance stop improving?}
In all the experiments, we have used a budget of 2\% for 5 rounds which corresponds to 10\% of the target dataset. We find that the performance of algorithm improves in majority of cycles.  This brings up the question, \textit{When does the performance of the model stop improving even after adding more labeled samples?}. For answering this question, we perform experiments on Webcam $\rightarrow$ Amazon and perform active DA for 30 rounds. Fig.~\ref{fig:conv} shows the results on Webcam $\rightarrow$ Amazon with S$^3$VAADA and Random sampling. It can be seen that after around 15 cycles, the gains due to additional samples being added decrease significantly and the performance seems to converge. The performance of the proposed S$^3$VAADA is much better than random sampling in all the rounds. It must also be noted that S$^3$VAADA reaches an accuracy of 89\% with 20 rounds (40\% of the dataset) which is equal to the performance when trained on all the target data.

\section{Analysis of VAADA training}
\label{improved_vada}

We propose VAADA method which is an enhanced version of VADA, suitable for Active DA. We find that proposed improvements in VAADA have a significant effect on the final active DA performance, which we analyse in detail in the following sections. We have done all our analysis using source dataset as Webcam and target dataset as Amazon which is a part of Office-31.
\subsection{Analysis of Learning Rate}
\begin{figure}[h]
  \centering
  \includegraphics[width=0.75\linewidth,height=5cm]{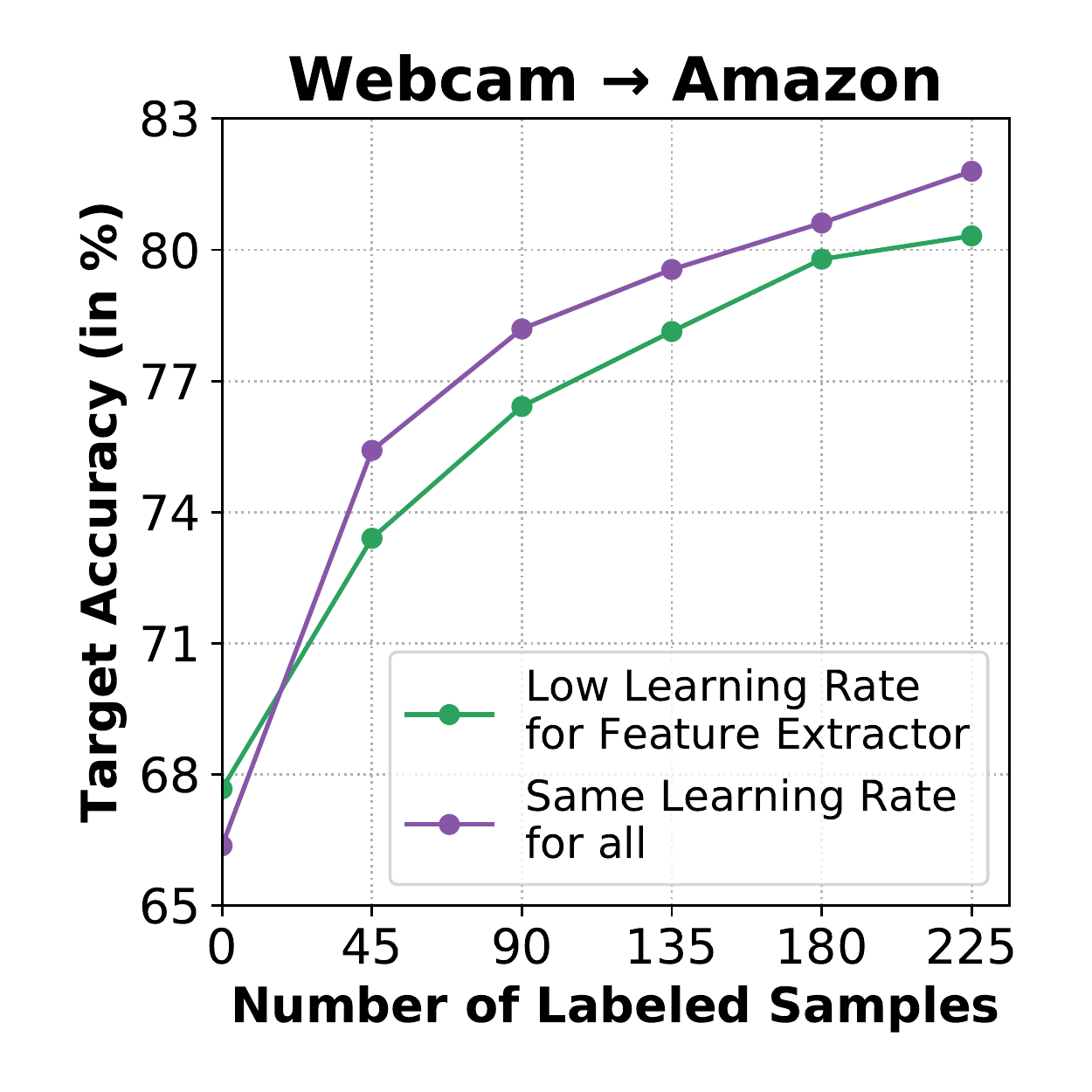}
  \caption{Comparison between Active DA with lower learning rate and a higher learning rate for backbone. The results are the average across three runs with different random seeds.}
  \label{fig:low-lr}
\end{figure}

It is a common practice ~\cite{ganin2015unsupervised, long2018conditional} in domain adaptation (DA) to use a relatively lower learning rate (usually decreased by a multiplying a factor of 0.1) for convolutional backbone which is ResNet-50 in our case. We find that though this practice helps for Unsupervised DA performance, it was not useful in the case of Active DA. In Fig.~\ref{fig:low-lr}, we show the comparison of using same learning rate for the backbone network (as proposed in VAADA), to using a smaller learning rate for backbone. The results clearly show that not lowering the learning is specially helpful for Active DA, whereas it is not for Unsupervised DA.

\subsection{Analysis of using Gradient Clipping}
\begin{figure}[h]
  \centering
  \includegraphics[width=0.75\linewidth,height=5cm]{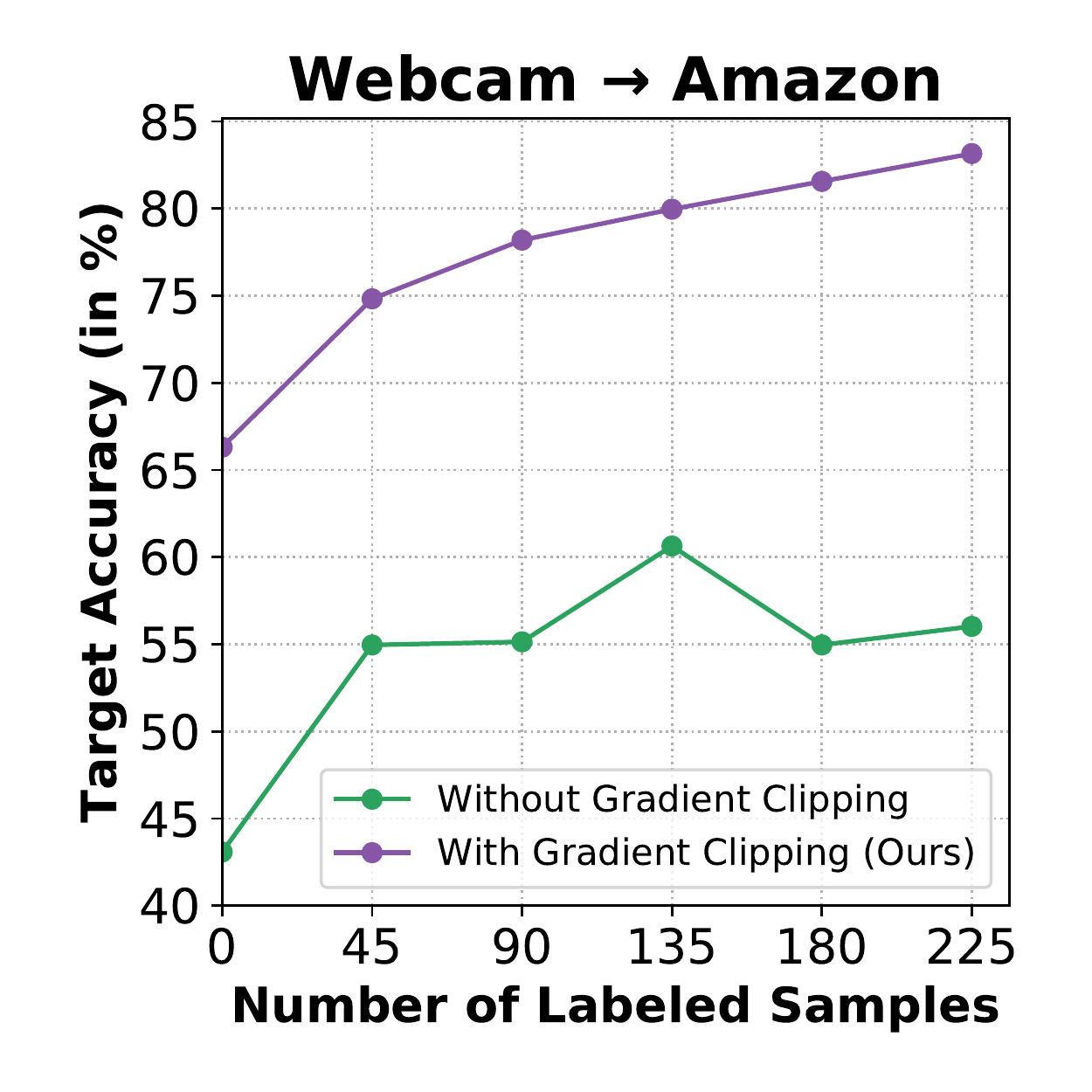}
  \caption{Ablating Gradient clipping on VAADA}
  \label{fig:no-gc}
\end{figure}

In the original implementation of VADA ~\cite{shu2018dirt} the authors use the method of Exponential Moving Average (EMA) (also known as Polyak Averaging ~\cite{polyak1992acceleration}) of model weights, which increases the stability of results. In place of EMA, we find that using proposed Gradient Clipping in VAADA works  better for stabilizing the training. In Gradient Clipping, we scale the gradients such that the gradient vector norm has magnitude 1. We find that Gradient Clipping allows the network to train stably, with a relatively high learning rate of 0.01. For showcasing the stabilising effect of Gradient Clipping, in Fig. \ref{fig:no-gc} we compare the performance of the model with and without gradient clipping. We find that Gradient Clipping leads to a increase of accuracy of above 10\% for each active learning cycle, with achieving stable increase in performance with the addition of more labels. On the other hand the model without clipping is unable to produce stable increase in performance with addition of labels. 

\subsection{Comparison of VADA with VAADA}
\begin{figure}[htp]
  \centering
  \includegraphics[width=\linewidth]{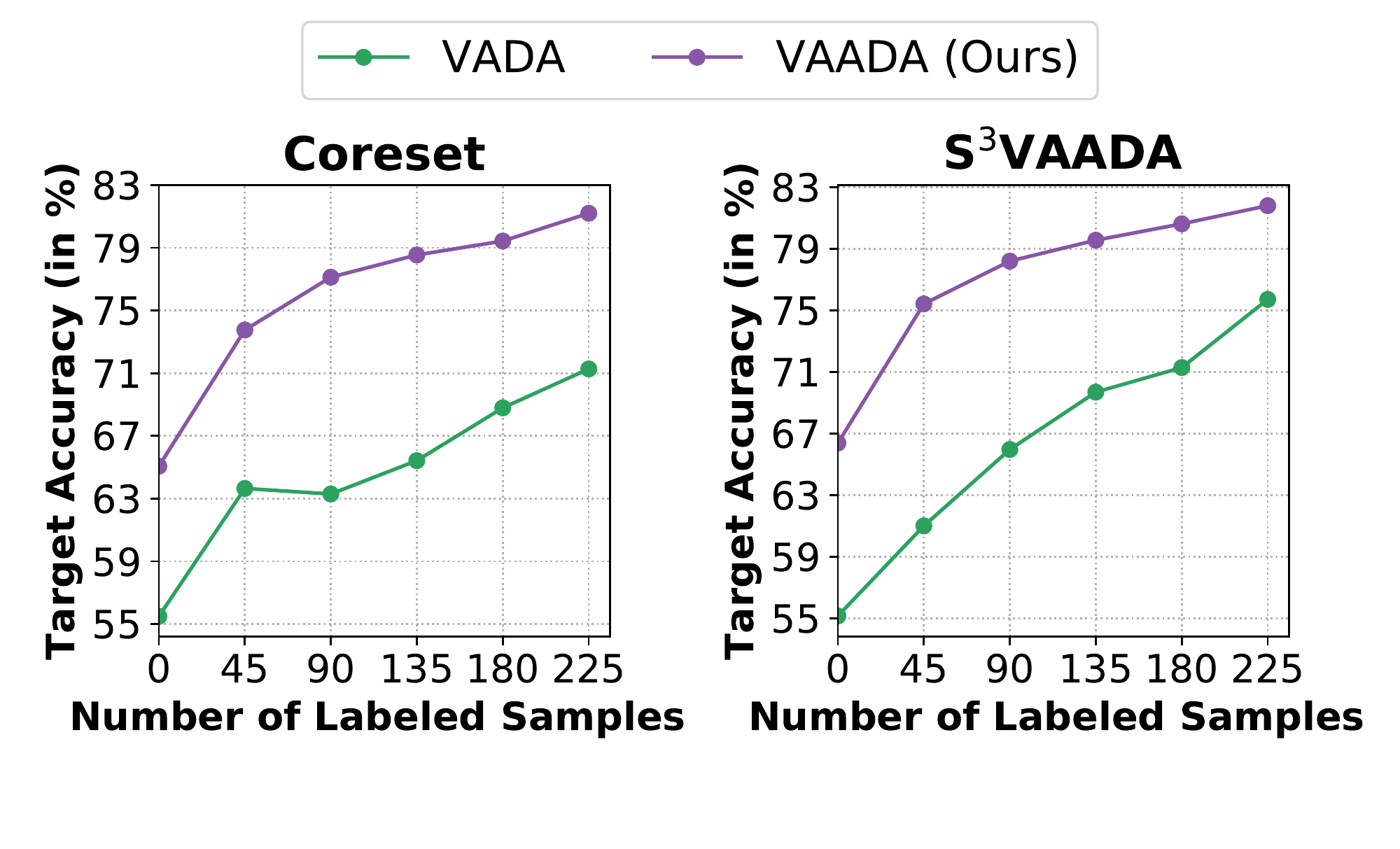}
  \vspace{-10mm}
  \caption{We show the comparison of VAADA and VADA. We see consistent improvement of VAADA over VADA  across all cycles.}
  \label{fig:ivada}
\end{figure}

In this section we provide additional implementation details and analysis, continuing from Sec. {\color{red} 6} of main paper. The comparison shown with the VADA method corresponds to the original VADA configuration specified in \cite{shu2018dirt}. In the original implementation, the authors propose to use Adam optimizer and EMA for training. We use Adam with learning rate of 0.0001 and use the exact same settings as in \cite{shu2018dirt}. It can be seen in Fig.~\ref{fig:ivada} that VAADA consistently outperforms the VADA training in Active DA for CoreSet and S$^3$VAADA as well.

\subsection{Visualizing clusters using t-SNE}
\begin{figure*}[h]
  \centering
  \subfigure[DANN]{\includegraphics[width=0.45\linewidth]{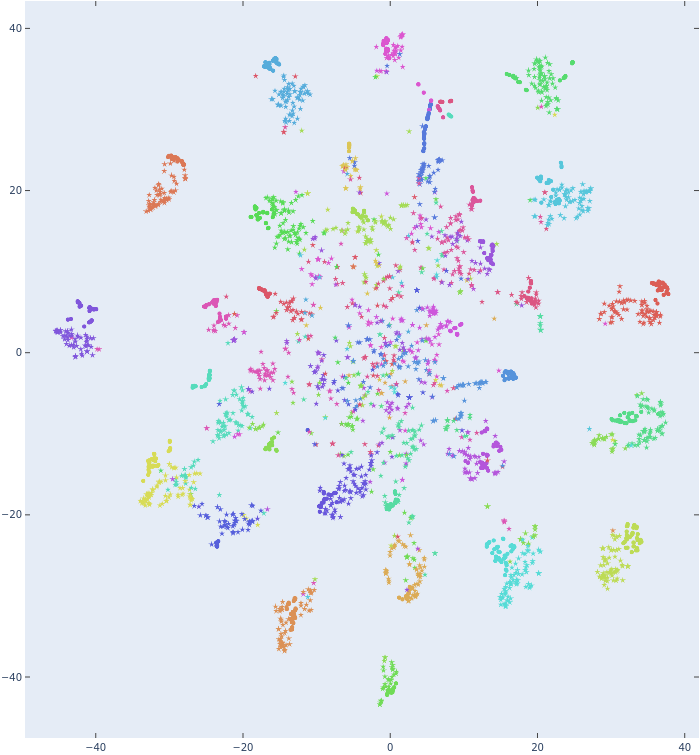}}
  \subfigure[VAADA]{\includegraphics[width=0.45\linewidth]{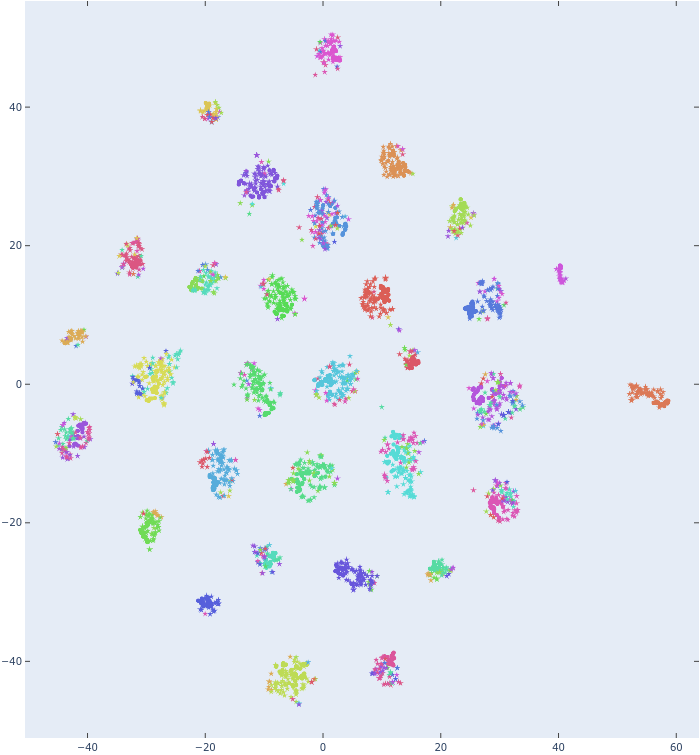}}
  \caption{Visualization of clusters of data points formed by DANN and VAADA on DA for Webcam $\rightarrow$ Amazon. Different colors represent different classes. It can seen that VAADA forms much distinct clusters data than DANN.}
  \label{fig:dann-vada}
\end{figure*}

In this section, we analyse the t-SNE plot (Fig.~\ref{fig:dann-vada}) of the two different training methods i.e., DANN and VAADA. We find that in VAADA training, there is formation of distinct clusters and also the cluster sizes are similar.  Whereas in DANN t-SNE, there is no formation of distinct clusters, and a large portion of sample are clustered in between. This shows that additional losses of conditional entropy and smoothing through Virtual Adversarial Perturbation loss are necessary to enforce the cluster assumption.

\subsection{Hyper-Parameter Sensitivity of VAADA}
\begin{figure}[h]
  \centering
  \includegraphics[width=0.75\linewidth,height=5cm]{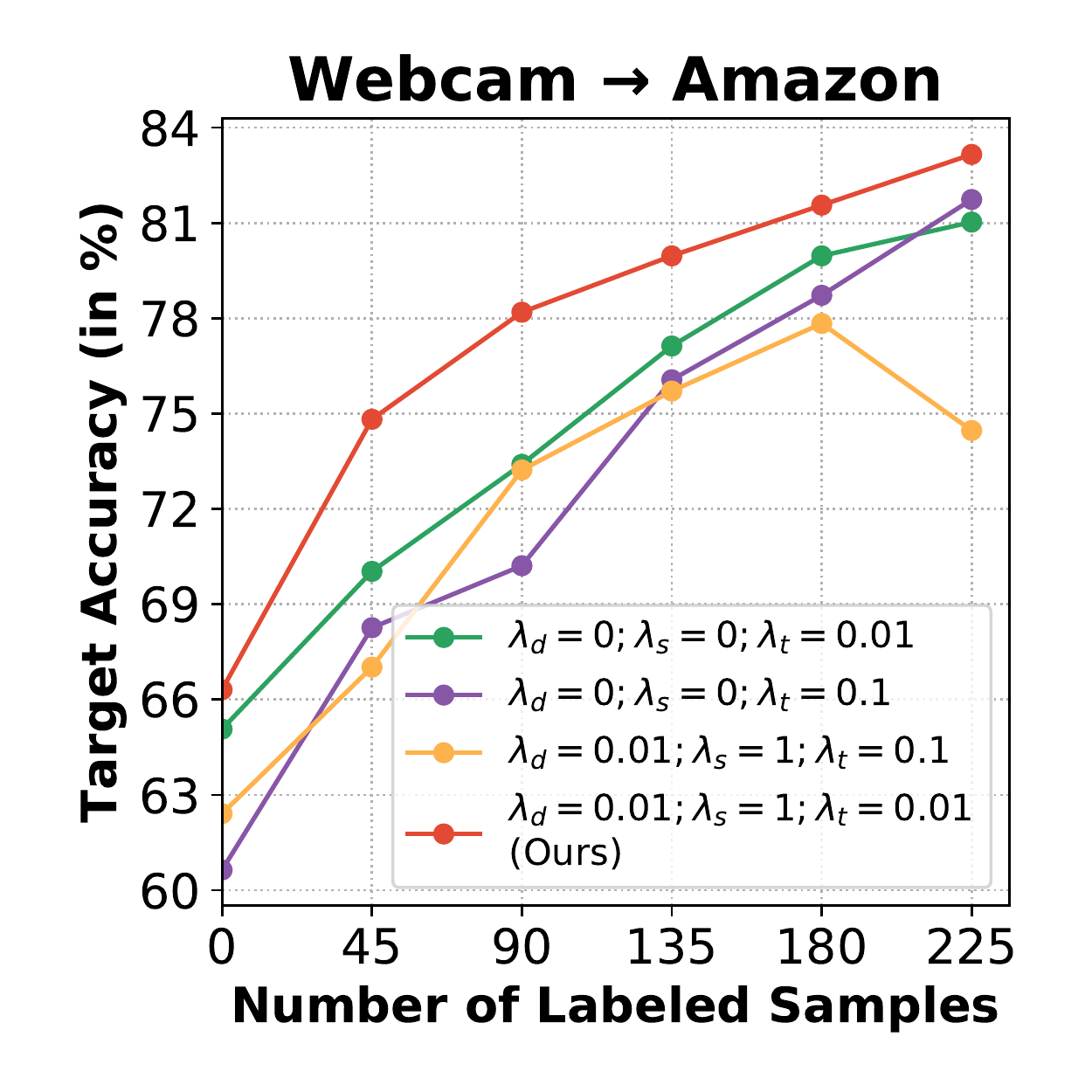}
  \caption{Different Hyperparameters on Webcam $\rightarrow$ Amazon dataset.}
  \label{fig:lambda-ablation}
\end{figure}

We used the same $\lambda$ values mentioned as a robust choice by VADA \cite{shu2018dirt} authors, for VAADA training, setting $\lambda_d = 0.01$, $\lambda_s = 1$ and $\lambda_t = 0.01$ across all datasets. For analysing the sensitivity of the performance of VAADA across different hyper-parameter choices, we provide results with varying $\lambda$ parameters in Fig. \ref{fig:lambda-ablation}. We also find that the robust choice recommended for VADA, also works the best for VAADA. Hence, this \textit{fixed-set} of robust $\lambda$ parameters can be used across datasets with varying degree of domain shifts. This is also enforced by the fact, that in all our experiments these \textit{fixed} hyperparameters were able to achieve state-of-the-art performance across datasets. This decreases the need for hyper-parameter tuning specific to each dataset. 

\section{Implementation Details}
\label{experimental_details}
\subsection{Configuration for DANN}
For the DANN experiments, we use a batch size of 36 with a learning rate of 0.01 for all the linear layers. We use a smaller learning rate of 0.001 for the ResNet-50 backbone. DANN is trained with SGD with a momentum of 0.9 and weight decay value of 0.0005 following the schedule described in \cite{ganin2015unsupervised}. The model architecture and hyperparameters are same as in \cite{long2018conditional}. The model is trained for 10,000 iterations as done in \cite{long2018conditional} and the best validation accuracy is reported in the graphs.
\subsection{Configuration for SSDA (MME*)}
We use the author's implementation\footnote{https://github.com/VisionLearningGroup/SSDA\_MME} for experiments on Office dataset. We used ResNet-50 as backbone and used same parameters as used in their implementation. For Active DA, we initially train the model with no labeled target data and keeps on adding 2\% of the unlabeled target data to labeled target set for 5 cycles. We train the model for 20,000 iterations. A similar procedure of reporting the best validation accuracy on the fixed validation set, as done for other baselines is followed.
\subsection{Configuration for VAADA}
The model is trained with a batch size of 16 and a learning rate of 0.01 for all the layers using the SGD Optimizer with a momentum of 0.9. A weight decay of 0.0005 was used. The model is trained for 100 epochs and the best accuracy is reported in the graphs. A ResNet-50 backbone is used with pretrained ImageNet weights. The architecture for various model components used are shown in Table \ref{tab:gen} and \ref{tab:clf}. Same architecture is used for all experiments in the paper. 

\begin{table}[]
    \centering
    \begin{tabular}{c||c}
    \hline
       Layer/Component  &  Output Shape\\
       \hline
        - & 224 $\times$ 224 $\times$ 3 \\
        ResNet-50 & 2048 \\
        Linear & 256 \\
        \hline
    \end{tabular}
    \caption{\textbf{Feature Generation $g_{\theta}$}: Architecture used for generating the features}
    \label{tab:gen}
\end{table}

\begin{table}[]
    \centering
    \begin{tabular}{c||c}
    \hline
       Layer  &  Output Shape\\
       \hline
       \multicolumn{2}{c}{\textbf{Feature Classifier ($f_{\theta}$)}} \\
       \hline
        - & 256 \\
        Linear & $C$ \\
        \hline
        \multicolumn{2}{c}{\textbf{Domain Classifier} ($D_\phi$)} \\
        \hline
        - & 256 \\
        Linear & 1024 \\
        ReLU & 1024 \\
        Linear & 1024 \\
        ReLU & 1024 \\
        Linear & 2 \\
        \hline
    \end{tabular}
    \caption{Architecture used for feature classifier and Domain classifier. $C$ is the number of classes. Both classifiers will take input from feature generator ($g_\theta$).}
    \label{tab:clf}
\end{table}

The above hyper parameters are used for all our experiments on Office-Home and Office-31 datasets. We just change the batch size to 128 and use the learning rate decay schedule of DANN for experiments on VisDA-18 dataset.

\begin{equation}
\label{eq:final-loss_supp}
\begin{split}
\begin{aligned}
                     L(\theta ; \mathcal{D}_{s}, \mathcal{D}_t, \mathcal{D}_u) = L_y(\theta; \mathcal{D}_s, \mathcal{D}_t) + \lambda_dL_d(\theta; \mathcal{D}_s, \mathcal{D}_t, \mathcal{D}_u) \notag\\ + \lambda_sL_{v}(\theta; \mathcal{D}_s \cup \mathcal{D}_t) + \lambda_t(L_v(\theta; \mathcal{D}_u) + L_c(\theta; \mathcal{D}_u)) \end{aligned}
\end{split}
\end{equation}

The $\epsilon$ used in Eq. 3 and 4 in the main paper refer to the maximum norm of the virtual adversarial perturbation,  was set it to 5 in our experiments.
The value of the number of random restarts ($N$) to generate virtual adversarial perturbation for the proposed sampling is set to 5. 
The $\alpha$ value is set to 0.5 and $\beta$ value is set to 0.3 across all experiments. We use Gradient Clipping to clip the norm of the gradient vector to 1 to stabilize and accelerate VAADA.
We used Weights \& Biases \cite{wandb} to track our experiments. 

\begin{figure}[h]
  \centering
  \includegraphics[width=0.75\linewidth,height=5cm]{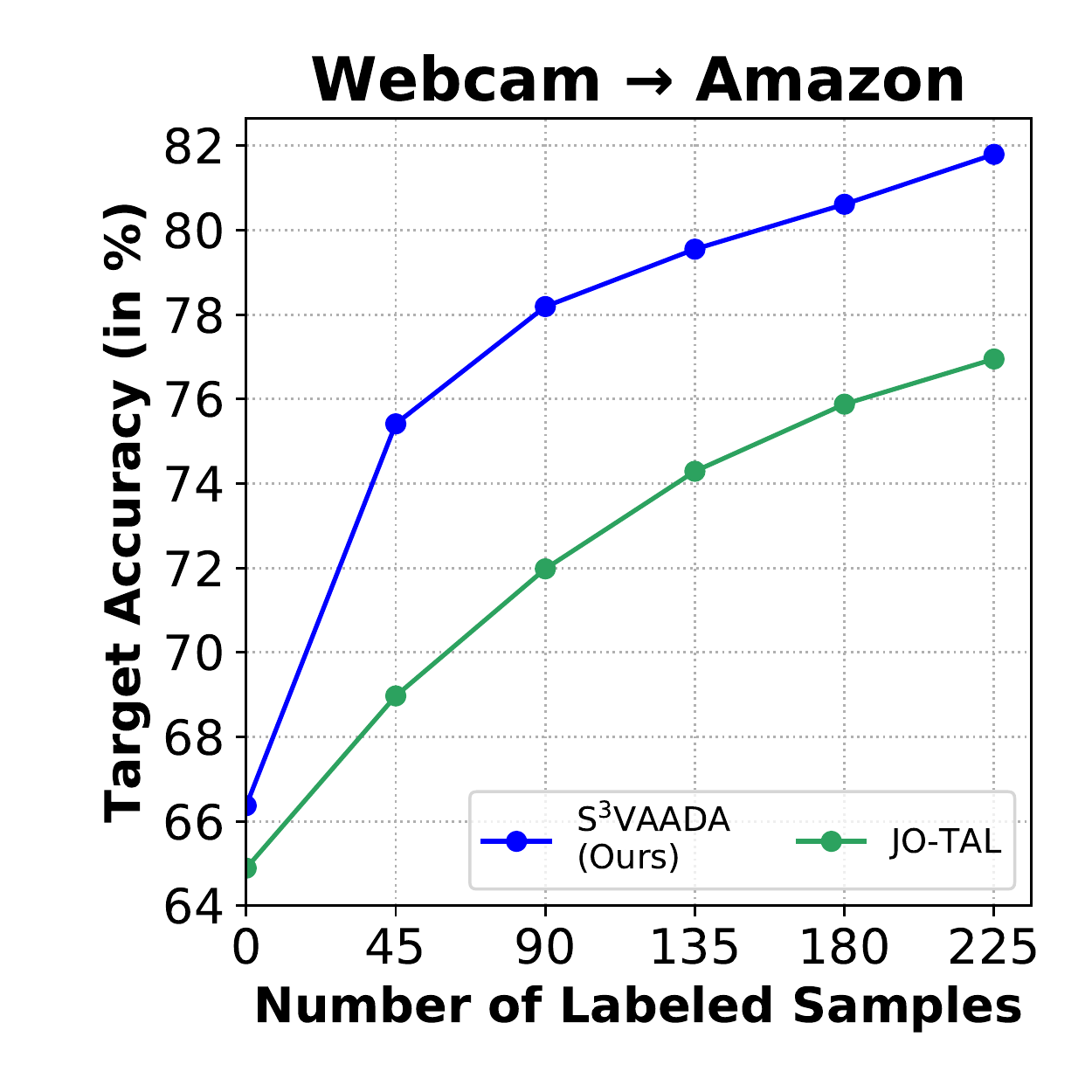}
  \caption{S$^3$VAADA vs. JO-TAL}
  \label{fig:jotal}
\end{figure}
\section{Comparison with JO-TAL}
\label{jotal}
We compare our results with JO-TAL by Chattopadhyay et al. \cite{chattopadhyay2013joint} (by implementing in \texttt{cvxopt}). JO-TAL performs both active learning and domain adaptation in a single step. Since, JO-TAL was not proposed in the context of deep learning, we use deep features from ImageNet pretraind model and train an SVM classifier on top of them. The optimization problem was implemented in \texttt{cvxopt}. Fig. \ref{fig:jotal} shows S$^3$VAADA achieves significant performance gains across cycles when compared to JO-TAL.

\section{Comparison with Alternate Adversarial Perturbation based sampling}
\label{Comparison with alternate Adversarial Sampling}
\begin{figure}[h]
  \centering
  \includegraphics[width=0.75\linewidth,height=5cm]{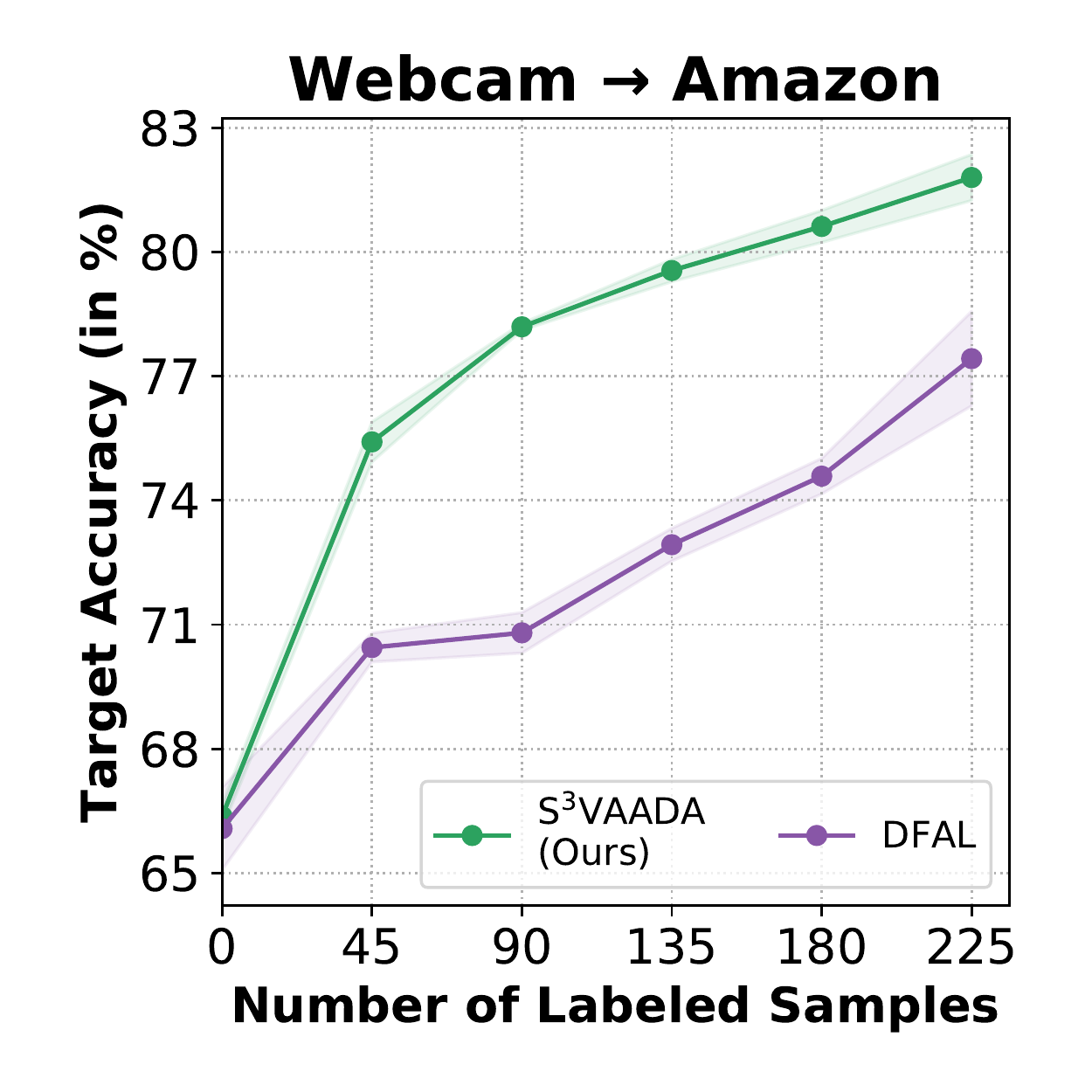}
  \caption{S$^3$VAADA outperforms DFAL in all the cycles, even though both attain same initial accuracy. It shows that S$^3$VAADA selects much more informative samples compared to DFAL.}
  \label{fig:dfal-svap}
\end{figure}

There also exists a sampling method \cite{ducoffe2018adversarial} based on DeepFool adversarial perturbations \cite{dezfooli2016deepfool} Active Learning (DFAL) but due to its higher complexity and computation time, it was unfeasible for us to use it as a baseline for all experiments. We provide the comparison of DFAL with S$^3$VAADA in terms of accuracy on Active DA from Webcam $\rightarrow$ Amazon in Fig. \ref{fig:dfal-svap}. Training is done through VAADA for both sampling methods. 
We find that S$^3$VAADA significantly outperforms DFAL sampling achieving better results in all cycles.

\begin{figure}[]
  \centering
  \subfigure[Webcam]{\fbox{\includegraphics[width=0.45\linewidth]{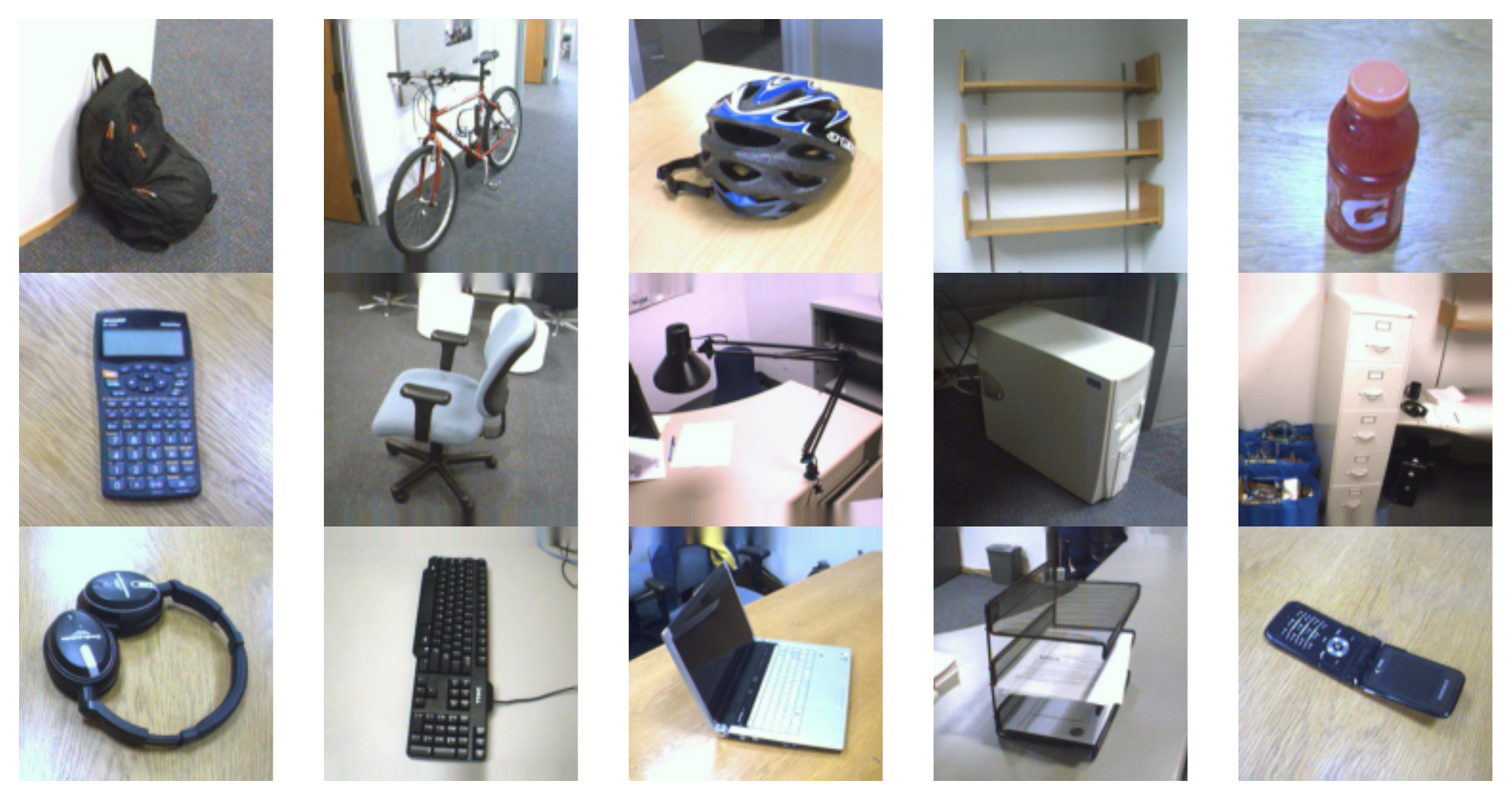}}}
  \subfigure[DSLR]{\fbox{\includegraphics[width=0.45\linewidth]{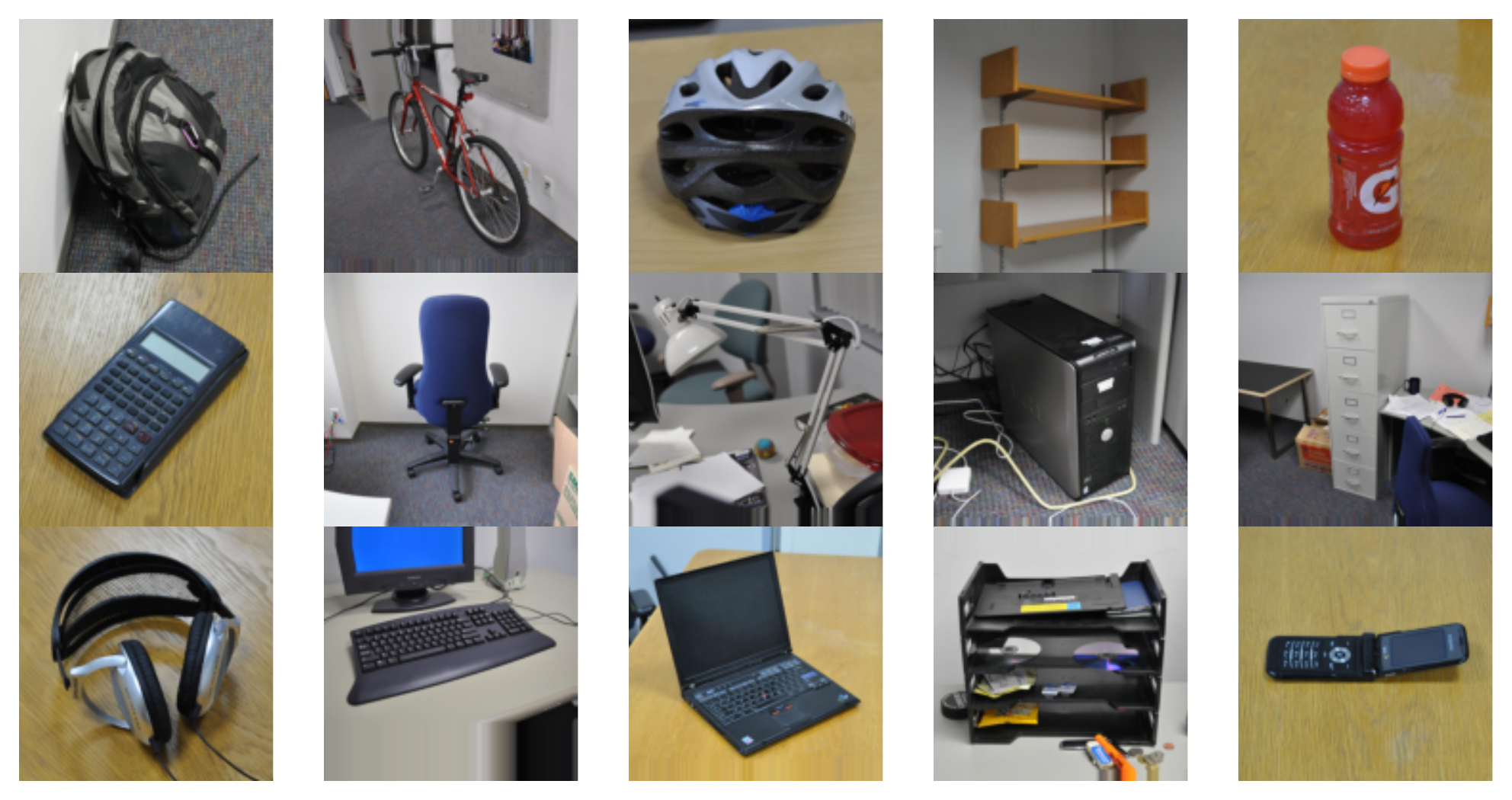}}}
  \subfigure[Amazon]{\fbox{\includegraphics[width=0.45\linewidth]{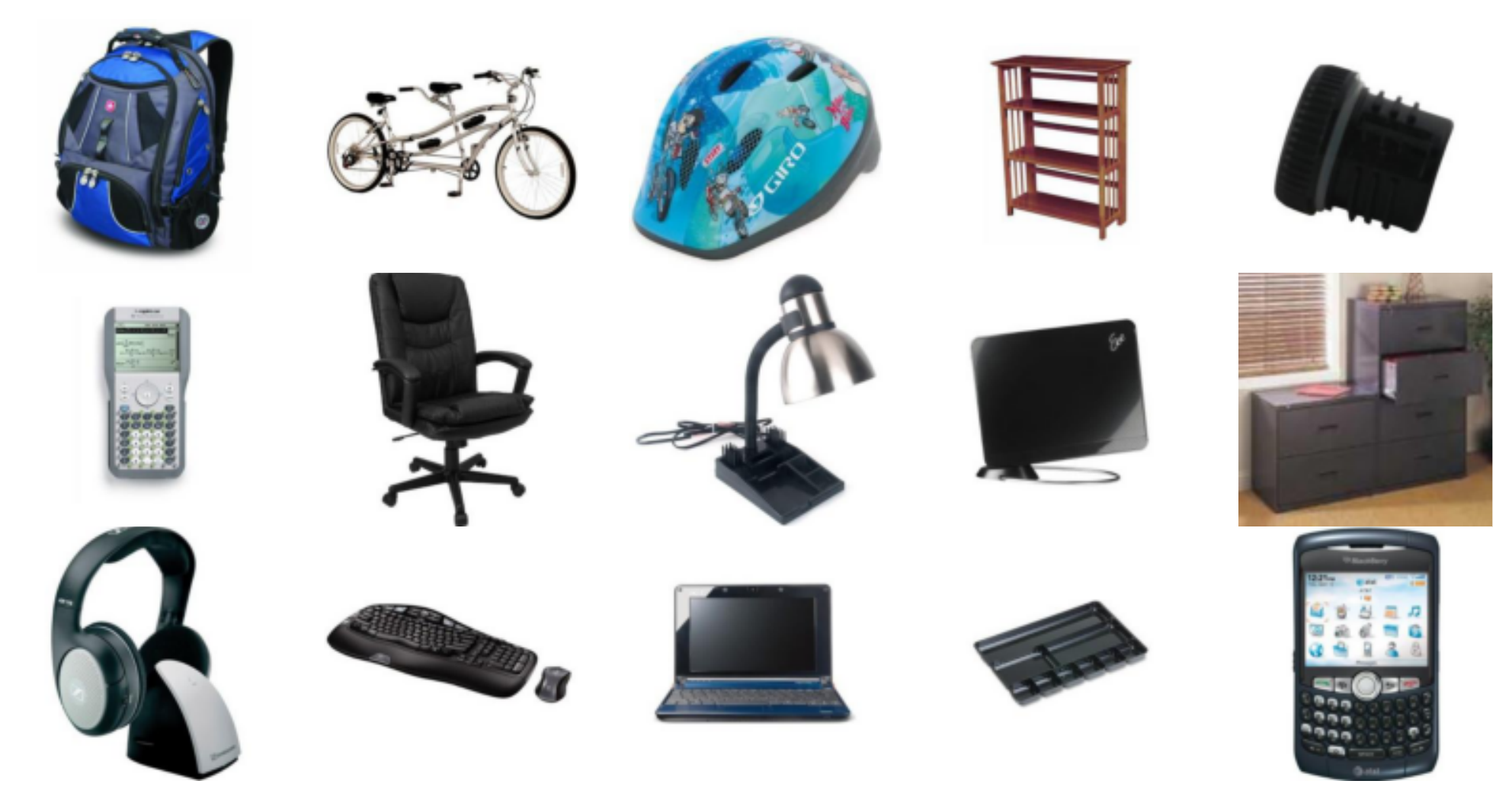}}}
  \caption{Some Office-31 Dataset examples}
  \label{fig:office-31-images}
\end{figure}

\begin{figure}[h]
  \centering
  \subfigure[Art]{\fbox{\includegraphics[width=0.45\linewidth]{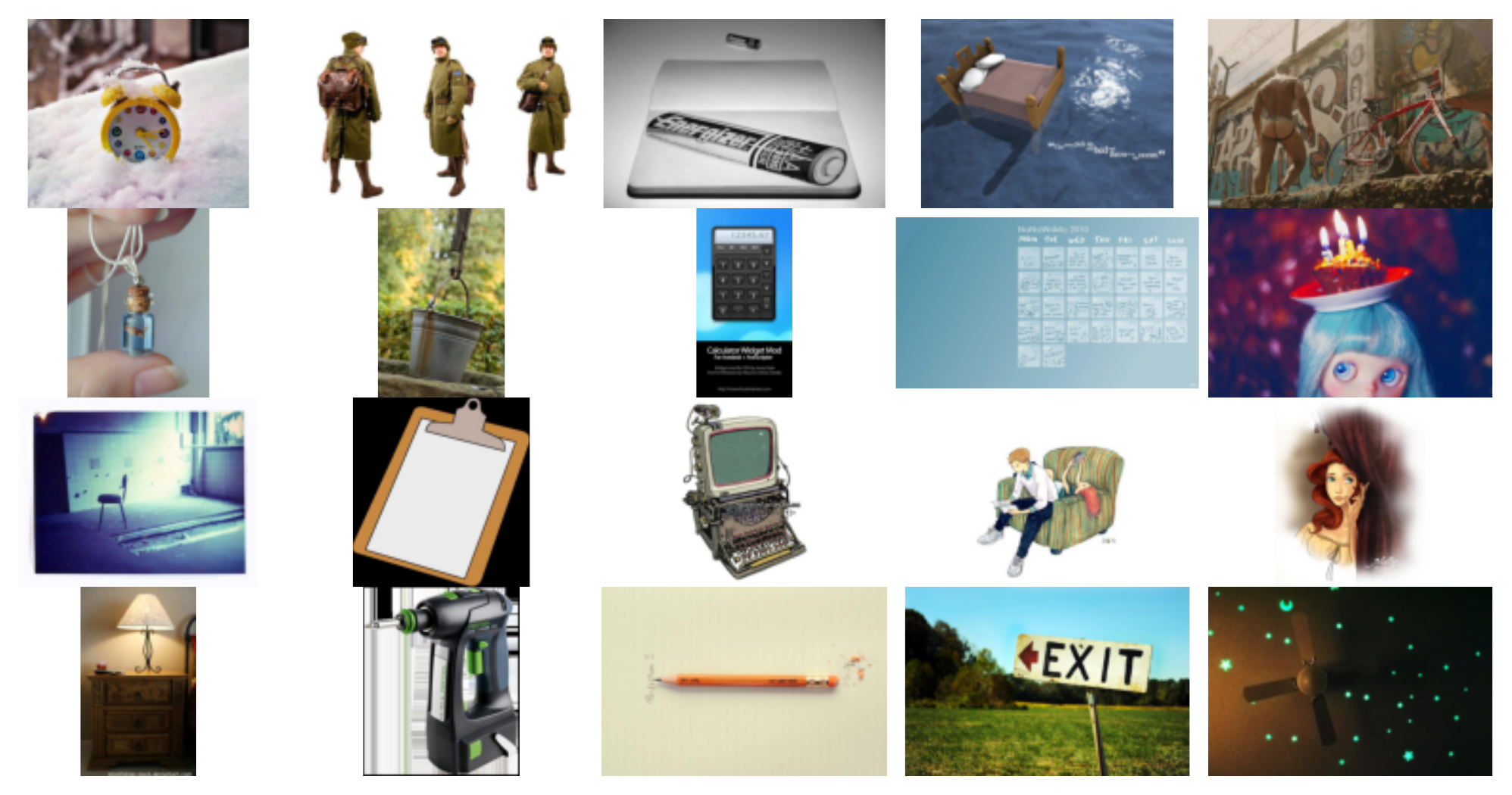}}}
  \subfigure[Product]{\fbox{\includegraphics[width=0.45\linewidth]{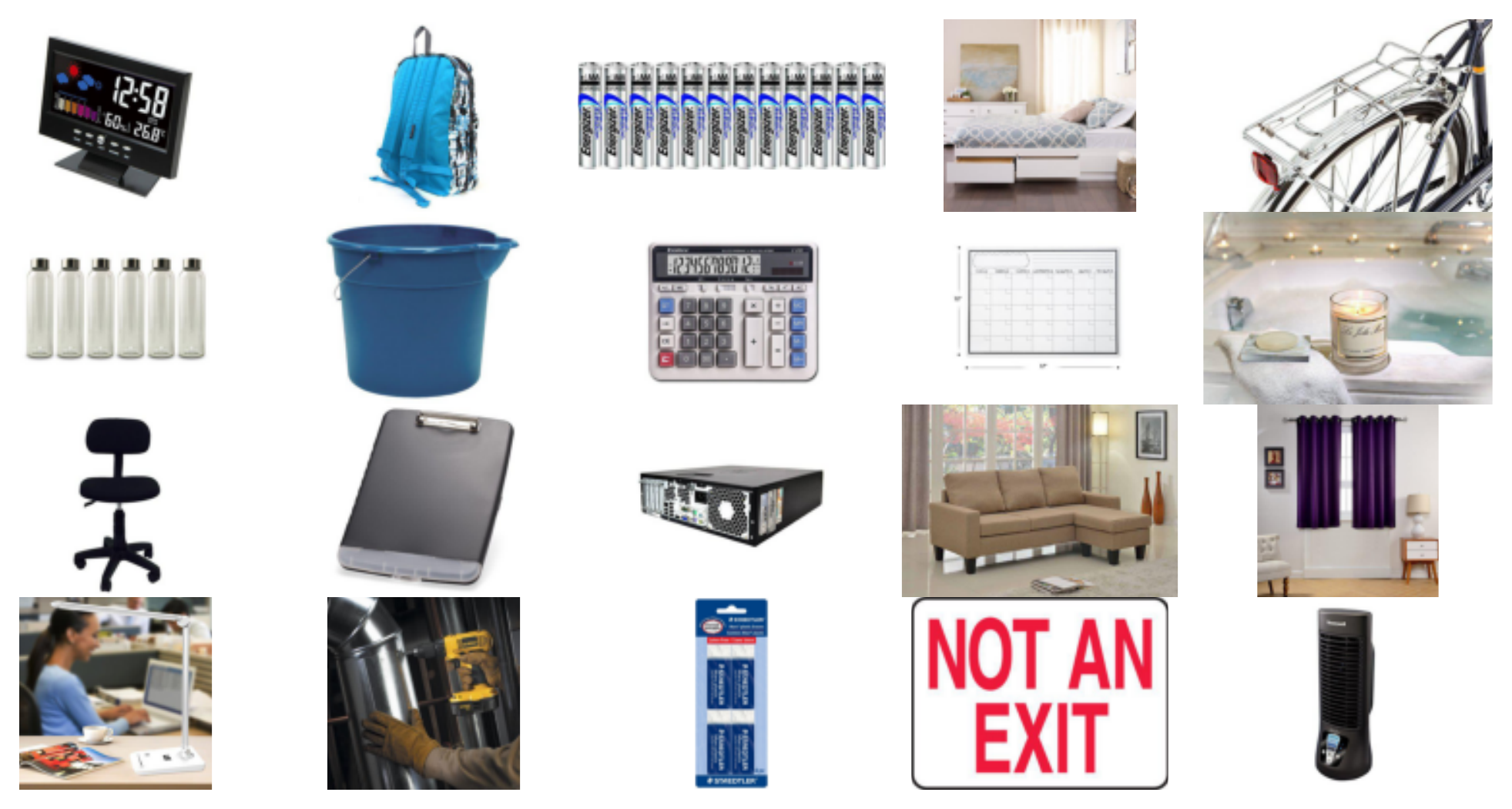}}}
  \subfigure[Clipart]{\fbox{\includegraphics[width=0.45\linewidth]{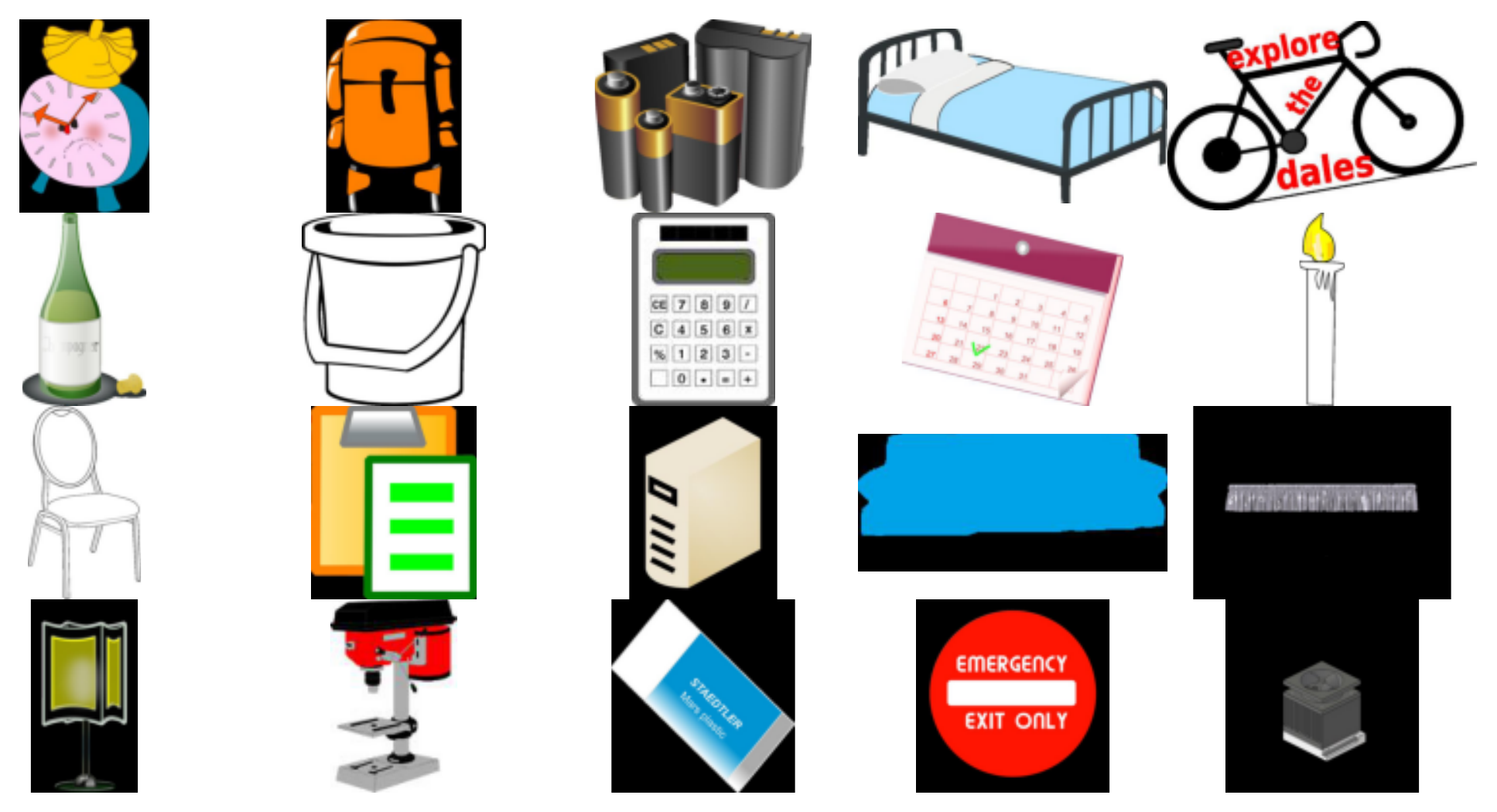}}}
  \subfigure[Real World]{\fbox{\includegraphics[width=0.45\linewidth]{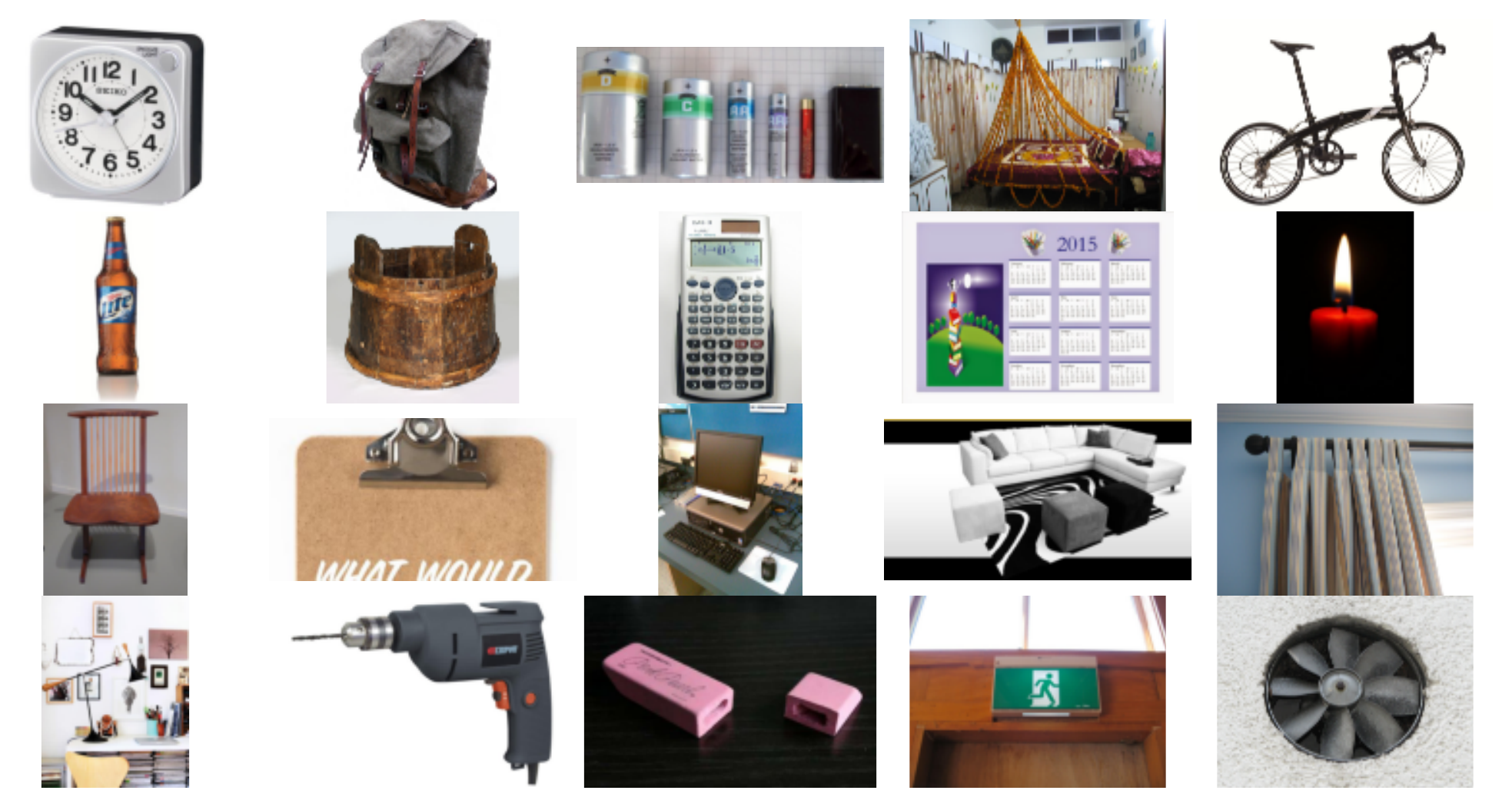}}}
  \caption{Some Office-Home Dataset examples}
  \label{fig:office-home-images}
\end{figure}

\begin{figure}[h]
  \centering
  \subfigure[Real]{\includegraphics[width=0.45\linewidth]{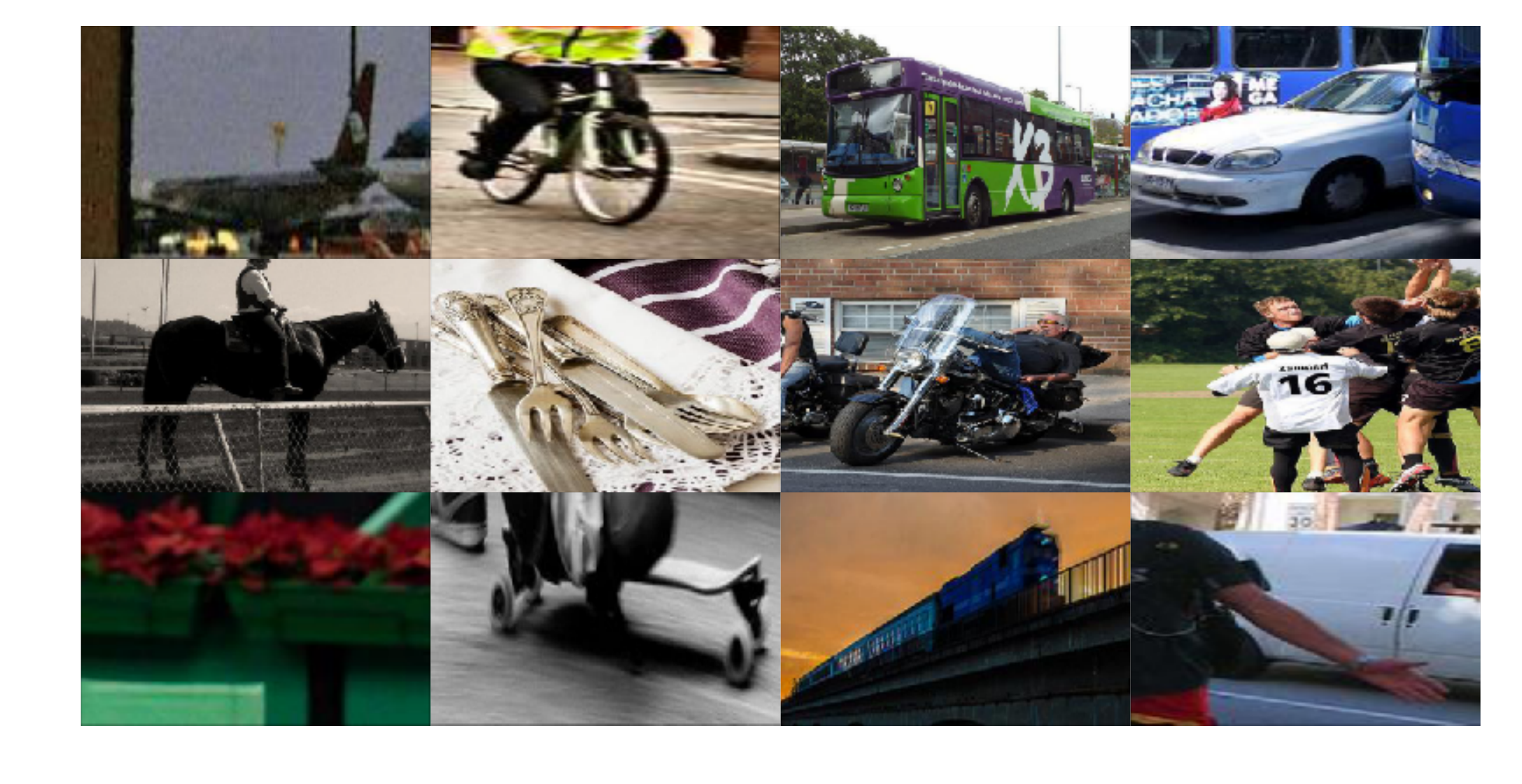}}
  \subfigure[Synthetic]{\includegraphics[width=0.45\linewidth]{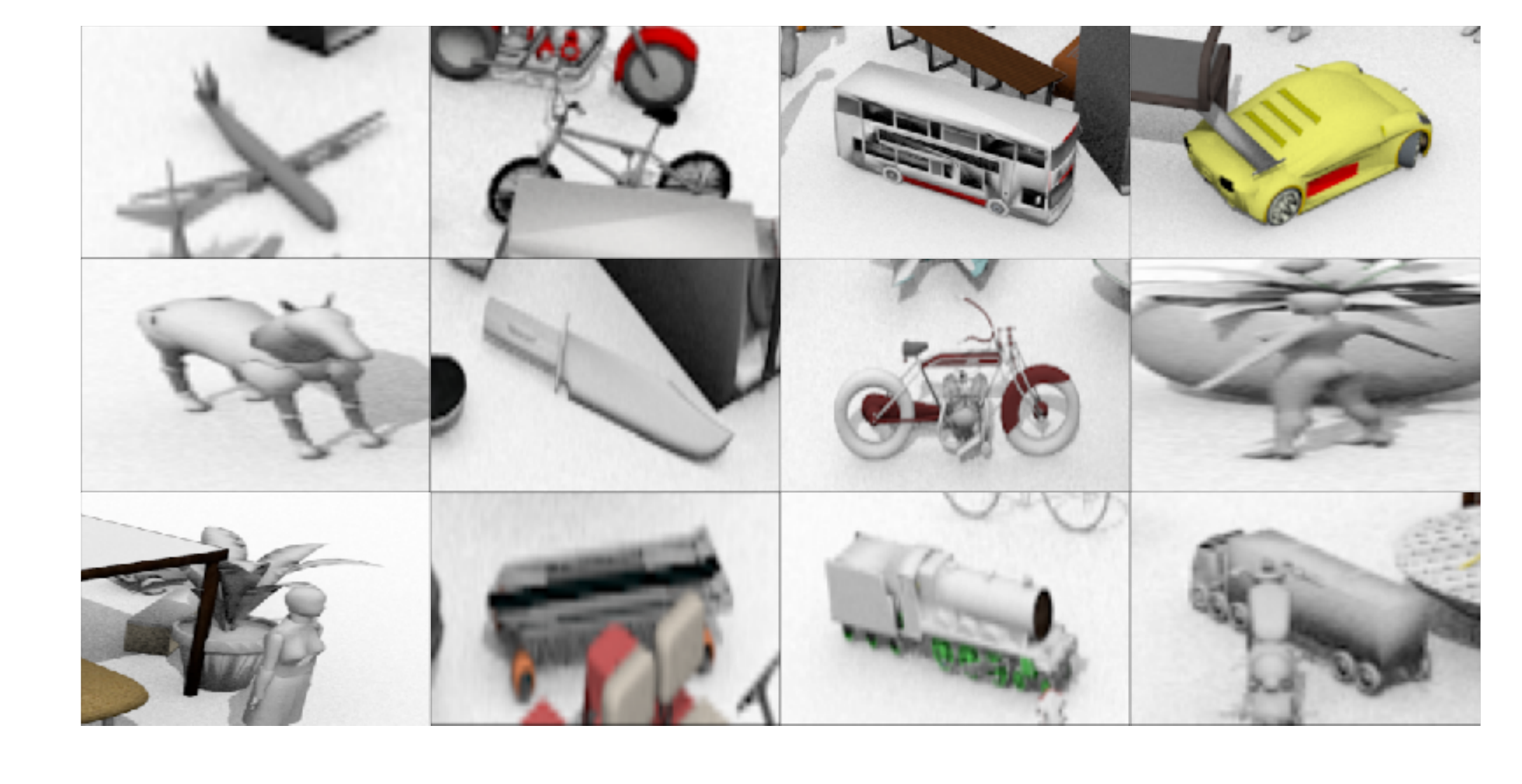}}
  \caption{Some Visual DA (VisDA-18) Dataset examples}
  \label{fig:visda-images}
\end{figure}

\section{Description of Datasets Used}
\label{datasets}
\textbf{Office-31 \cite{saenko2010adapting}:} It has images from 3 domains i.e., Webcam, DSLR and Amazon, belonging to 31 classes.

\textbf{Office-Home \cite{venkateswara2017Deep}:} This dataset has a more severe domain shift across domains  compared to Office-31. It is a 65 class dataset and contains images from 4 domains namely, Art, Clipart, Product and Real World. 

\textbf{VisDA-18 \cite{8575439}:} This dataset consist of images from synthetic and real domains. The dataset has annotations for for two tasks: image classification and image segmentation. We used dataset of image classification task. It has 12 different object categories.

Some example images of each dataset are shown in Figs. \ref{fig:office-31-images}, \ref{fig:office-home-images} and \ref{fig:visda-images}.

\begin{figure}[htp]
  \centering
  \includegraphics[width=0.75\linewidth,height=5cm]{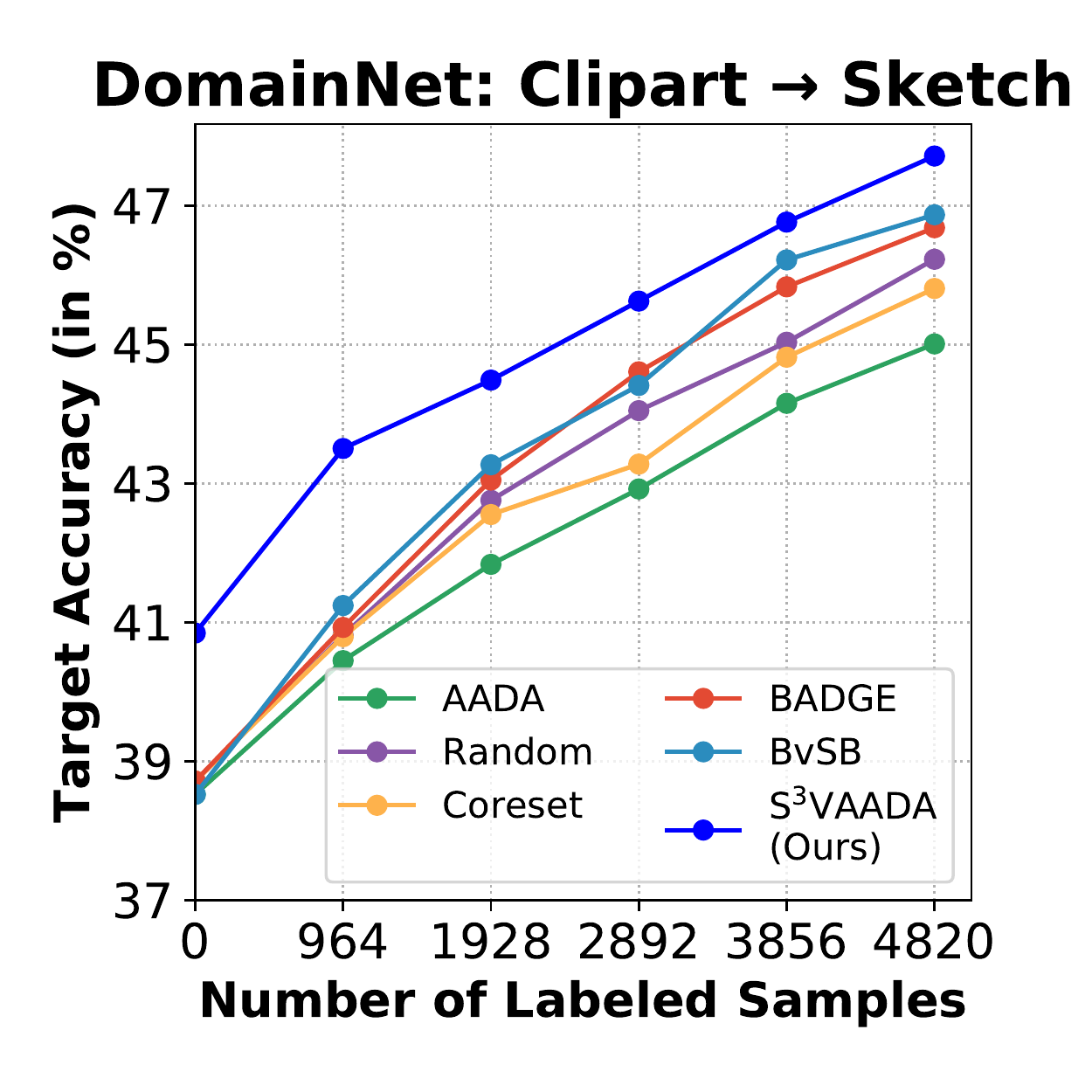}
  \caption{Active Domain Adaptation results on Clipart $\rightarrow$ Sketch dataset. This shows the proposed method is scalable to larger datasets.}
  \label{fig:c2s}
\end{figure}

\section{DomainNet Experiments}
\label{domainnet}
DomainNet \cite{peng2019moment} consists of about 0.6 million images belonging to 345 classes. The images belong to 5 domains: Clipart, Sketch, Quickdraw, Painting, Real. For showing the scalability of our method, we use Clipart as the source and Sketch as the target domain. The Clipart domain consists of 33,525 images in the train set and 14,604 images in the test set. The Sketch domain consists of 50,416 images in the train set and 21,850 images in the test set. Due to computational limitations we are not able to provide results on different possible domains.

For the DomainNet experiments, we use a batch size of 36 with a learning rate scheduler same as DANN and run each baseline for 30 epochs. We use Gradient Clipping and clip the norm to 10. In this case we find that performance of S$^3$VAADA does not stagnate at 30 epochs but due to limitations of compute we only train for 30 epochs. Hence, there exist scope for improvement in results with parameter tuning and more computational budget.

Fig.~\ref{fig:c2s} shows the results on Clipart $\rightarrow$ Sketch domain shift. The performance of S$^3$VAADA outperforms all the other techniques in all the cycles. This shows that the efficacy of the proposed method on a large dataset containing 345 classes. 

\section{Future Extension to Other Applications}
The S$^3$VAADA technique is based on the idea of cluster assumption i.e., aligning of clusters of different classes, which is used in the sampling method. Some recent DA techniques for Object Detection \cite{xu2020cross} and Image Segmentation \cite{wang2020classes} which aim for classwise alignment of features, can be seen as methods which satisfy the cluster assumption. Hence, we hope that combining such techniques with our method can yield good Active DA techniques tailored for these specific applications. In the current work, we focused on diverse image classification tasks, leaving these applications for future work.

{\small
\bibliographystyle{ieee_fullname}
\bibliography{egbib}

\begin{thebibliography}{10}\itemsep=-1pt

\bibitem{Ash2020Deep}
Jordan~T. Ash, Chicheng Zhang, Akshay Krishnamurthy, John Langford, and Alekh
  Agarwal.
\newblock Deep batch active learning by diverse, uncertain gradient lower
  bounds.
\newblock In {\em International Conference on Learning Representations}, 2020.

\bibitem{10.2307/25047882}
A. Bhattacharyya.
\newblock On a measure of divergence between two multinomial populations.
\newblock {\em Sankhyā: The Indian Journal of Statistics (1933-1960)},
  7(4):401--406, 1946.

\bibitem{wandb}
Lukas Biewald.
\newblock Experiment tracking with weights and biases, 2020.
\newblock Software available from wandb.com.

\bibitem{chakraborty2014adaptive}
Shayok Chakraborty, Vineeth Balasubramanian, and Sethuraman Panchanathan.
\newblock Adaptive batch mode active learning.
\newblock {\em IEEE transactions on neural networks and learning systems},
  26(8):1747--1760, 2014.

\bibitem{chattopadhyay2013joint}
Rita Chattopadhyay, Wei Fan, Ian Davidson, Sethuraman Panchanathan, and Jieping
  Ye.
\newblock Joint transfer and batch-mode active learning.
\newblock In {\em International conference on machine learning}, pages
  253--261. PMLR, 2013.

\bibitem{chen2020adversarial}
Minghao Chen, Shuai Zhao, Haifeng Liu, and Deng Cai.
\newblock Adversarial-learned loss for domain adaptation.
\newblock In {\em Proceedings of the AAAI Conference on Artificial
  Intelligence}, volume~34, pages 3521--3528, 2020.

\bibitem{chen2020adversariallearned}
Minghao Chen, Shuai Zhao, Haifeng Liu, and Deng Cai.
\newblock Adversarial-learned loss for domain adaptation.
\newblock {\em arXiv}, abs/2001.01046, 2020.

\bibitem{cohn1994improving}
David Cohn, Les Atlas, and Richard Ladner.
\newblock Improving generalization with active learning.
\newblock {\em Machine learning}, 15(2):201--221, 1994.

\bibitem{deng2019cluster}
Zhijie Deng, Yucen Luo, and Jun Zhu.
\newblock Cluster alignment with a teacher for unsupervised domain adaptation.
\newblock In {\em Proceedings of the IEEE/CVF International Conference on
  Computer Vision}, pages 9944--9953, 2019.

\bibitem{ducoffe2018adversarial}
Melanie Ducoffe and Frederic Precioso.
\newblock Adversarial active learning for deep networks: a margin based
  approach.
\newblock {\em arXiv preprint arXiv:1802.09841}, 2018.

\bibitem{ganin2015unsupervised}
Yaroslav Ganin and Victor Lempitsky.
\newblock Unsupervised domain adaptation by backpropagation.
\newblock In {\em International conference on machine learning}, pages
  1180--1189. PMLR, 2015.

\bibitem{Goluba2000EigenvalueCI}
H. Goluba and Henk~A. van~der Vorstb.
\newblock Eigenvalue computation in the 20 th century gene.
\newblock 2000.

\bibitem{He_2016_CVPR}
Kaiming He, Xiangyu Zhang, Shaoqing Ren, and Jian Sun.
\newblock Deep residual learning for image recognition.
\newblock In {\em Proceedings of the IEEE Conference on Computer Vision and
  Pattern Recognition (CVPR)}, June 2016.

\bibitem{joseph2019submodular}
KJ Joseph, Krishnakant Singh, Vineeth~N Balasubramanian, et~al.
\newblock Submodular batch selection for training deep neural networks.
\newblock {\em arXiv preprint arXiv:1906.08771}, 2019.

\bibitem{5206627}
A.~J. {Joshi}, F. {Porikli}, and N. {Papanikolopoulos}.
\newblock Multi-class active learning for image classification.
\newblock In {\em 2009 IEEE Conference on Computer Vision and Pattern
  Recognition}, pages 2372--2379, 2009.

\bibitem{krause2009optimizing}
Andreas Krause and Carlos Guestrin.
\newblock Optimizing sensing: From water to the web.
\newblock {\em Computer}, 42(8):38--45, 2009.

\bibitem{Kundu_2020_CVPR}
Jogendra~Nath Kundu, Naveen Venkat, Ambareesh Revanur, Rahul~M V, and
  R.~Venkatesh Babu.
\newblock Towards inheritable models for open-set domain adaptation.
\newblock In {\em Proceedings of the IEEE/CVF Conference on Computer Vision and
  Pattern Recognition (CVPR)}, June 2020.

\bibitem{Kundu_2020_CVPR_usfda}
Jogendra~Nath Kundu, Naveen Venkat, Rahul~M V, and R.~Venkatesh Babu.
\newblock Universal source-free domain adaptation.
\newblock In {\em Proceedings of the IEEE/CVF Conference on Computer Vision and
  Pattern Recognition (CVPR)}, June 2020.

\bibitem{lee2019drop}
Seungmin Lee, Dongwan Kim, Namil Kim, and Seong-Gyun Jeong.
\newblock Drop to adapt: Learning discriminative features for unsupervised
  domain adaptation.
\newblock In {\em Proceedings of the IEEE/CVF International Conference on
  Computer Vision}, pages 91--100, 2019.

\bibitem{li2020online}
Da Li and Timothy Hospedales.
\newblock Online meta-learning for multi-source and semi-supervised domain
  adaptation.
\newblock In {\em European Conference on Computer Vision}, pages 382--403.
  Springer, 2020.

\bibitem{long2018conditional}
Mingsheng Long, Zhangjie Cao, Jianmin Wang, and Michael~I Jordan.
\newblock Conditional adversarial domain adaptation.
\newblock In {\em Advances in Neural Information Processing Systems}, pages
  1645--1655, 2018.

\bibitem{miyato2019vat}
T. {Miyato}, S. {Maeda}, M. {Koyama}, and S. {Ishii}.
\newblock Virtual adversarial training: A regularization method for supervised
  and semi-supervised learning.
\newblock {\em IEEE Transactions on Pattern Analysis and Machine Intelligence},
  41(8):1979--1993, 2019.

\bibitem{dezfooli2016deepfool}
S. {Moosavi-Dezfooli}, A. {Fawzi}, and P. {Frossard}.
\newblock Deepfool: A simple and accurate method to fool deep neural networks.
\newblock In {\em 2016 IEEE Conference on Computer Vision and Pattern
  Recognition (CVPR)}, pages 2574--2582, 2016.

\bibitem{nemhauser1978analysis}
George~L Nemhauser, Laurence~A Wolsey, and Marshall~L Fisher.
\newblock An analysis of approximations for maximizing submodular set
  functions—i.
\newblock {\em Mathematical programming}, 14(1):265--294, 1978.

\bibitem{NEURIPS2019_9015}
Adam Paszke, Sam Gross, Francisco Massa, Adam Lerer, James Bradbury, Gregory
  Chanan, Trevor Killeen, Zeming Lin, Natalia Gimelshein, Luca Antiga, Alban
  Desmaison, Andreas Kopf, Edward Yang, Zachary DeVito, Martin Raison, Alykhan
  Tejani, Sasank Chilamkurthy, Benoit Steiner, Lu Fang, Junjie Bai, and Soumith
  Chintala.
\newblock Pytorch: An imperative style, high-performance deep learning library.
\newblock In H. Wallach, H. Larochelle, A. Beygelzimer, F. d\textquotesingle
  Alch\'{e}-Buc, E. Fox, and R. Garnett, editors, {\em Advances in Neural
  Information Processing Systems 32}, pages 8024--8035. Curran Associates,
  Inc., 2019.

\bibitem{peng2019moment}
Xingchao Peng, Qinxun Bai, Xide Xia, Zijun Huang, Kate Saenko, and Bo Wang.
\newblock Moment matching for multi-source domain adaptation.
\newblock In {\em Proceedings of the IEEE International Conference on Computer
  Vision}, pages 1406--1415, 2019.

\bibitem{8575439}
X. {Peng}, B. {Usman}, N. {Kaushik}, D. {Wang}, J. {Hoffman}, and K. {Saenko}.
\newblock Visda: A synthetic-to-real benchmark for visual domain adaptation.
\newblock In {\em 2018 IEEE/CVF Conference on Computer Vision and Pattern
  Recognition Workshops (CVPRW)}, pages 2102--21025, 2018.

\bibitem{polyak1992acceleration}
Boris~T Polyak and Anatoli~B Juditsky.
\newblock Acceleration of stochastic approximation by averaging.
\newblock {\em SIAM journal on control and optimization}, 30(4):838--855, 1992.

\bibitem{prabhu2020active}
Viraj Prabhu, Arjun Chandrasekaran, Kate Saenko, and Judy Hoffman.
\newblock Active domain adaptation via clustering uncertainty-weighted
  embeddings, 2020.

\bibitem{rai2010domain}
Piyush Rai, Avishek Saha, Hal Daum{\'e}~III, and Suresh Venkatasubramanian.
\newblock Domain adaptation meets active learning.
\newblock In {\em Proceedings of the NAACL HLT 2010 Workshop on Active Learning
  for Natural Language Processing}, pages 27--32, 2010.

\bibitem{saenko2010adapting}
Kate Saenko, Brian Kulis, Mario Fritz, and Trevor Darrell.
\newblock Adapting visual category models to new domains.
\newblock In {\em European conference on computer vision}, pages 213--226.
  Springer, 2010.

\bibitem{saito2019semi}
Kuniaki Saito, Donghyun Kim, Stan Sclaroff, Trevor Darrell, and Kate Saenko.
\newblock Semi-supervised domain adaptation via minimax entropy.
\newblock In {\em Proceedings of the IEEE International Conference on Computer
  Vision}, pages 8050--8058, 2019.

\bibitem{saito2018adversarial}
Kuniaki Saito, Yoshitaka Ushiku, Tatsuya Harada, and Kate Saenko.
\newblock Adversarial dropout regularization.
\newblock In {\em International Conference on Learning Representations}, 2018.

\bibitem{saito2018maximum}
Kuniaki Saito, Kohei Watanabe, Yoshitaka Ushiku, and Tatsuya Harada.
\newblock Maximum classifier discrepancy for unsupervised domain adaptation.
\newblock In {\em Proceedings of the IEEE Conference on Computer Vision and
  Pattern Recognition}, pages 3723--3732, 2018.

\bibitem{sener2018active}
Ozan Sener and Silvio Savarese.
\newblock Active learning for convolutional neural networks: A core-set
  approach.
\newblock In {\em International Conference on Learning Representations}, 2018.

\bibitem{sener2016learning}
Ozan Sener, Hyun~Oh Song, Ashutosh Saxena, and Silvio Savarese.
\newblock Learning transferrable representations for unsupervised domain
  adaptation.
\newblock In {\em Advances in Neural Information Processing Systems}, pages
  2110--2118, 2016.

\bibitem{settles2009active}
Burr Settles.
\newblock Active learning literature survey.
\newblock 2009.

\bibitem{shu2018dirt}
Rui Shu, Hung Bui, Hirokazu Narui, and Stefano Ermon.
\newblock A dirt-t approach to unsupervised domain adaptation.
\newblock In {\em International Conference on Learning Representations}, 2018.

\bibitem{sinha2019variational}
Samarth Sinha, Sayna Ebrahimi, and Trevor Darrell.
\newblock Variational adversarial active learning.
\newblock In {\em Proceedings of the IEEE/CVF International Conference on
  Computer Vision}, pages 5972--5981, 2019.

\bibitem{Su_2020_WACV}
Jong-Chyi Su, Yi-Hsuan Tsai, Kihyuk Sohn, Buyu Liu, Subhransu Maji, and
  Manmohan Chandraker.
\newblock Active adversarial domain adaptation.
\newblock In {\em Proceedings of the IEEE/CVF Winter Conference on Applications
  of Computer Vision (WACV)}, March 2020.

\bibitem{tong2001support}
Simon Tong and Daphne Koller.
\newblock Support vector machine active learning with applications to text
  classification.
\newblock {\em Journal of machine learning research}, 2(Nov):45--66, 2001.

\bibitem{tsai2018learning}
Yi-Hsuan Tsai, Wei-Chih Hung, Samuel Schulter, Kihyuk Sohn, Ming-Hsuan Yang,
  and Manmohan Chandraker.
\newblock Learning to adapt structured output space for semantic segmentation.
\newblock In {\em Proceedings of the IEEE conference on computer vision and
  pattern recognition}, pages 7472--7481, 2018.

\bibitem{vassilvitskii2006k}
Sergei Vassilvitskii and David Arthur.
\newblock k-means++: The advantages of careful seeding.
\newblock In {\em Proceedings of the eighteenth annual ACM-SIAM symposium on
  Discrete algorithms}, pages 1027--1035, 2006.

\bibitem{venkateswara2017Deep}
Hemanth Venkateswara, Jose Eusebio, Shayok Chakraborty, and Sethuraman
  Panchanathan.
\newblock Deep hashing network for unsupervised domain adaptation.
\newblock In {\em ({IEEE}) Conference on Computer Vision and Pattern
  Recognition ({CVPR})}, 2017.

\bibitem{wang2014new}
Dan Wang and Yi Shang.
\newblock A new active labeling method for deep learning.
\newblock In {\em 2014 International joint conference on neural networks
  (IJCNN)}, pages 112--119. IEEE, 2014.

\bibitem{wang2020classes}
Haoran Wang, Tong Shen, Wei Zhang, Ling-Yu Duan, and Tao Mei.
\newblock Classes matter: A fine-grained adversarial approach to cross-domain
  semantic segmentation.
\newblock In {\em European Conference on Computer Vision}, pages 642--659.
  Springer, 2020.

\bibitem{enwiki:1012600046}
{Wikipedia contributors}.
\newblock Facility location problem --- {Wikipedia}{,} the free encyclopedia,
  2021.
\newblock [Online; accessed 17-March-2021].

\bibitem{xu2020cross}
Minghao Xu, Hang Wang, Bingbing Ni, Qi Tian, and Wenjun Zhang.
\newblock Cross-domain detection via graph-induced prototype alignment.
\newblock In {\em Proceedings of the IEEE/CVF Conference on Computer Vision and
  Pattern Recognition}, pages 12355--12364, 2020.

\bibitem{yoo2019learning}
Donggeun Yoo and In~So Kweon.
\newblock Learning loss for active learning.
\newblock In {\em Proceedings of the IEEE/CVF Conference on Computer Vision and
  Pattern Recognition}, pages 93--102, 2019.

\end{thebibliography}
}

\end{document}